\documentclass[10pt,journal,compsoc]{IEEEtran}

\ifCLASSOPTIONcompsoc
  \usepackage[nocompress]{cite}
\else
  \usepackage{cite}
\fi

\ifCLASSINFOpdf
\else
\fi

\usepackage{amsmath,amssymb} 
\usepackage{amsfonts}
\usepackage{graphicx}
\usepackage{comment}
\usepackage{tabularx}
\usepackage{booktabs}
\usepackage{subfigure}
\usepackage{wrapfig}
\usepackage{multirow}
\usepackage{arydshln}
\usepackage{color}
\usepackage{hyperref}
\usepackage{url}
\usepackage{xspace}
\usepackage{xcolor}
\usepackage{bm}

\def\h{\mathbf{h}}
\def\n{\mathbf{n}}

\def\x{\mathbf{x}}
\def\y{\mathbf{y}}
\def\z{\mathbf{z}}

\def\H{\mathbf{H}}

\def\W{\mathbf{W}}

\makeatletter
\DeclareRobustCommand\onedot{\futurelet\@let@token\@onedot}
\def\@onedot{\ifx\@let@token.\else.\null\fi\xspace}
 
\def\eg{\emph{e.g}\onedot} 
\def\ie{\emph{i.e}\onedot}

\def\wrt{w.r.t\onedot} 
\def\aka{\emph{a.k.a}\onedot}
\def\etal{\emph{et al}\onedot}
\makeatother

\DeclareMathOperator*{\argmax}{arg\,max}

\newcolumntype{Y}{>{\centering\arraybackslash}X}

\begin{document}

\title{Learning Latent Part-Whole Hierarchies \\ for Point Clouds}
\author{Xiang~Gao,~\IEEEmembership{Student Member,~IEEE,}
        Wei~Hu,~\IEEEmembership{Senior~Member,~IEEE,}
        and~Renjie~Liao
\IEEEcompsocitemizethanks{
\IEEEcompsocthanksitem X. Gao and W. Hu are with Wangxuan Institute of Computer Technology, Peking University, No. 128 Zhongguancun North Street, Beijing, China.\protect\\
E-mail: \{gyshgx868, forhuwei\}@pku.edu.cn
\IEEEcompsocthanksitem R. Liao is with Department of Electrical and Computer Engineering, University of British Columbia, Vancouver, Canada. E-mail: rjliao@ece.ubc.ca
\IEEEcompsocthanksitem Corresponding author: Wei Hu. 
}
}

\markboth{Journal of \LaTeX\ Class Files,~Vol.~14, No.~8, August~2015}%
{Shell \MakeLowercase{\textit{et al.}}: Bare Demo of IEEEtran.cls for Computer Society Journals}
%

\IEEEtitleabstractindextext{%
\begin{abstract}
Strong evidence suggests that humans perceive the 3D world by parsing visual scenes and objects into part-whole hierarchies. 
Although deep neural networks have the capability of learning powerful multi-level representations, they can not explicitly model part-whole hierarchies, which limits their expressiveness and interpretability in processing 3D vision data such as point clouds.
To this end, we propose an encoder-decoder style latent variable model that explicitly learns the part-whole hierarchies for the multi-level point cloud segmentation. 
Specifically, the encoder takes a point cloud as input and predicts the per-point latent subpart distribution at the middle level.
The decoder takes the latent variable and the feature from the encoder as an input and predicts the per-point part distribution at the top level.
During training, only annotated part labels at the top level are provided, thus making the whole framework weakly supervised. 
We explore two kinds of approximated inference algorithms, \ie, most-probable-latent and Monte Carlo methods, and three stochastic gradient estimations for learning discrete latent variables, \ie, straight-through, REINFORCE, and pathwise estimators.
Experimental results on the PartNet dataset show that the proposed method achieves state-of-the-art performance in not only top-level part segmentation but also middle-level latent subpart segmentation.
\end{abstract}

}

\maketitle

\IEEEdisplaynontitleabstractindextext
\IEEEpeerreviewmaketitle

\IEEEraisesectionheading{\section{Introduction}
\label{sec:intro}}

Research works in psychology \cite{biederman1987recognition,wolfs1994parallel,tanaka2016parts}, cognitive science \cite{hoffman1984parts,kinzler2007core}, and neural science \cite{hochstein2002view,peelen2007neural} suggest that humans perceive objects in a visual scene as part-whole hierarchies.
Albeit being simple to human visual systems, the visual reasoning of part-whole hierarchies has been a long-standing challenge for machine learning models, especially for neural networks \cite{hinton1979some,hinton1990mapping,riesenhuber1996neural}.
The deep learning breakthrough enables impressive progress in learning useful visual representations at multiple levels.
However, the multi-scale feature hierarchy within a deep neural network does not explicitly correspond to the part-whole hierarchy, thus leading to a few recent novel architecture designs that aim at solving this problem for 2D images \cite{hinton2021represent,sun2021visual,garau2022interpretable}.

\begin{figure}
  \centering
  \includegraphics[width=\columnwidth]{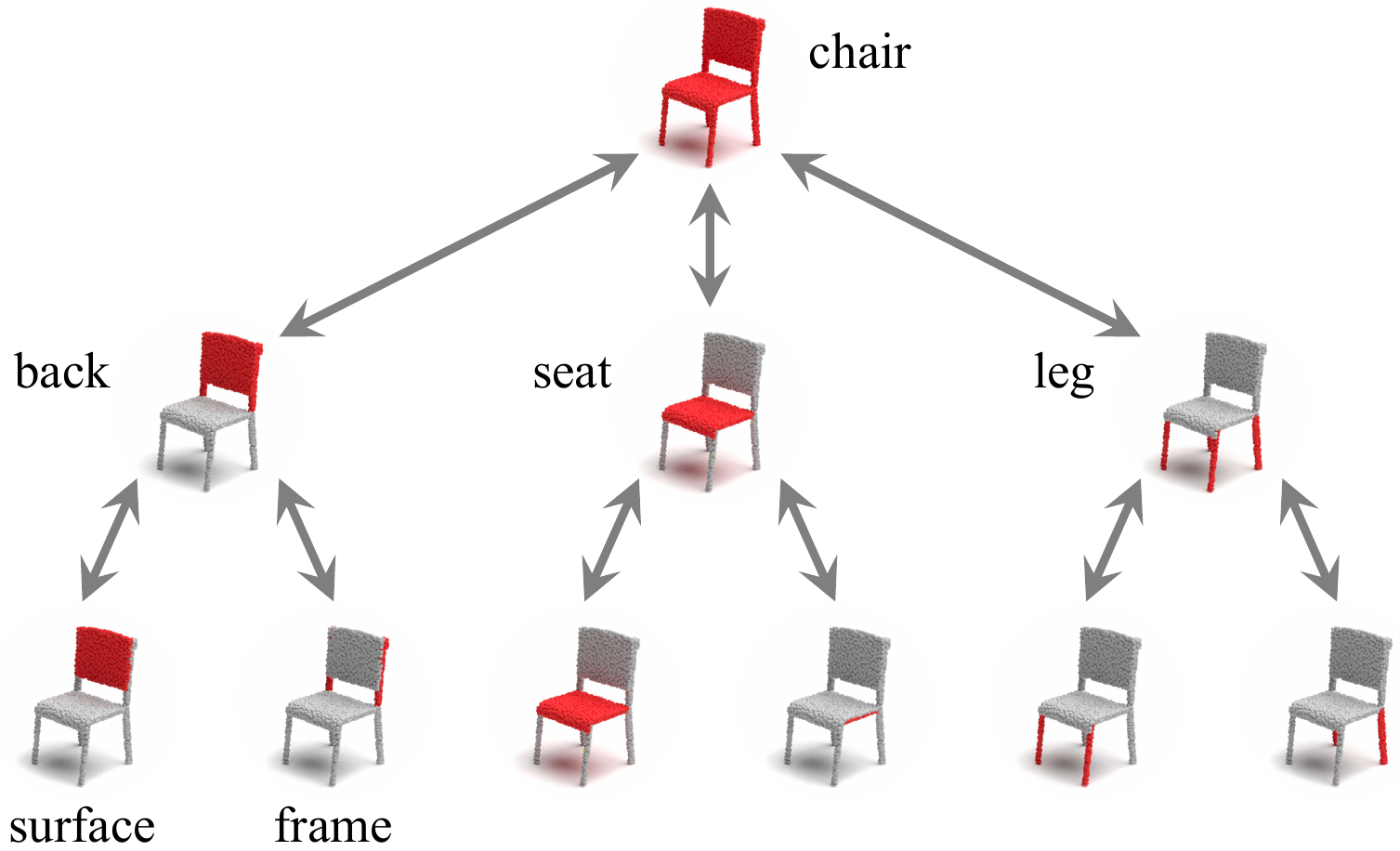}
  \caption{\textbf{An illustration of the part-whole hierarchies for 3D objects.} A chair can be decomposed into backs, legs, seats in a higher level, and each back may consist of frames and surfaces at a lower level. 
  }
  \label{fig:teaser}
\end{figure}

However, few attempts have been made for 3D point clouds---a set of points irregularly sampled from continuous surfaces that represent 3D objects or scenes. 
With the maturity of techniques like the depth sensing, the laser scanning and the image processing, point clouds become widely available in applications including autonomous driving, robotics, and immersive telepresence.  
Compared to 2D images, point clouds exhibit abundant 3D structural information of an object or a scene. 
As shown in Fig.~\ref{fig:teaser}, objects often have multiple levels of categories, \eg, a chair can be composed of backs, legs, seats, and other components at a higher level, while each back may consist of frames and surfaces at a lower level.
Explicitly modeling of such part-whole hierarchies for point clouds is challenging for the following reasons.
First, it is extremely expensive to annotate semantic labels of individual points within part-whole hierarchies, thus making the powerful supervised learning paradigm impractical.
The more parts a 3D object has or the deeper a part-whole hierarchy is, the more annotation efforts are required.
Second, due to the fact that point clouds are a set of irregularly sampled points, one needs to build perception models that are permutation invariant and 3D-geometry aware.
This introduces more constraints in modeling compared to 2D images.
Third, the more parts at different levels, the more expressive a part-whole hierarchy is and the more complicated the inference and learning algorithms become.
It is thus challenging to design efficient inference and learning that can achieve the best trade-off between expressiveness and tractability. 

In this paper, we propose a discrete latent variable model that learns part-whole hierarchies for point clouds in a weakly-supervised manner. 
Given a set of point clouds, we assume some annotated labels at some higher levels of the part-whole hierarchy are given, \eg, the class label of the whole set of point clouds or the part labels at the highest possible level.
We treat all part or sub-part labels of individual points at remaining lower levels of the part-whole hierarchy as categorical latent variables.
For example, for point cloud segmentation problems in our experiments, we consider two levels, \ie, annotated part labels are given in the top level whereas sub-part labels in the bottom level are latent.
We consider these two levels since current public datasets only provide annotations for these two levels so that we can quantitatively evaluate our latent variable models. 
Nevertheless, as discussed in Section~\ref{subsec:discuss}, our model can be generalized to more levels in a straightforward way. 

We design an encoder that takes raw point clouds as input and predicts the conditional probability distribution of these latent variables.
We then draw samples (\ie, latent sub-part labels in the bottom level) from the conditional distribution and feed them to a decoder that predicts the top-level part labels of the input point cloud.
As shown later, our modeling technique is general in that the latent variable module can be easily plugged into popular deep learning models to learn meaningful latent random variables in other domains/tasks. 
Since the state space of all latent variables is huge, approximated Bayesian inference methods are needed.
Even worse, our latent variables are discrete, thus disabling the popular differentiable reparameterization tricks (\eg, those used in VAEs \cite{kingma2013auto}) for gradient based learning.
Therefore, we systematically investigate the most-probable and Monte Carlo methods for approximate inference and straight-through \cite{bengio2013estimating}, REINFORCE \cite{williams1992simple}, and relaxed pathwise (\eg, Gumbel-softmax \cite{jang2016categorical,maddison2016concrete}) gradient estimators for learning.
The whole framework falls under the weakly-supervised learning paradigm as we have supervised annotations only at higher levels but not lower levels.
To summarize, our main contributions are as follows.
\begin{itemize}
  \item We propose a discrete latent variable model based on an encoder-decoder style neural network that can learn part-whole hierarchies for point clouds in a weakly-supervised manner, \eg, supervision is only provided on the top-level rather than the middle level of the hierarchy.
  \item We systematically investigate approximated inference methods, including the most-probable-latent and Monte Carlo methods, and stochastic gradient based learning methods for discrete latent variables, including the straight-through, REINFORCE, and the pathwise gradient estimators.
  \item Experiments on weakly-supervised multi-level point cloud segmentation datasets demonstrate that the proposed method not only achieves better top-level performances but also outperforms existing methods in learning part-whole hierarchies.
\end{itemize}

The remaining paper is organized as follows.
We first review related works in Section~\ref{sec:related_works}.
Then we introduce the proposed latent variable model in Section~\ref{sec:model}.
The inference and learning algorithms are described in Section~\ref{sec:inference} and Section~\ref{sec:learning}, respectively.
Finally, we show experimental results in Section~\ref{sec:experiments} and conclude the paper in Section~\ref{sec:conclusion}.

\section{Related Works}
\label{sec:related_works}
In this section, we review previous works on learning part-whole hierarchies in computer vision and deep learning methods for 3D point clouds.

\subsection{Learning Part-Whole Hierarchies}

One of the defining features of deep neural networks is that they can learn powerful hierarchical/multi-level representations.
Especially in computer vision, hierarchical representation learning is baked into architectures like convolutional neural networks (CNNs), \eg, \cite{ronneberger2015u}, \cite{lin2017feature}, \cite{zhao2017pyramid}.
Bottom layers of CNNs typically learn spatially local (low-level) features such as textures and oriented edges, while top layers learn spatially global (high-level) features that have more semantic meanings such as object classes.

However, common neural networks do not explicitly learn part-whole hierarchies since the implicit hierarchical representation learning does not provide strong enough semantic information to indicate where various parts and wholes are.
Many attempts have been made to design machine learning models to explicitly learn part-whole hierarchies in images.
Hinton \cite{hinton1990mapping} propose three types of neural networks that can achieve this goal and discuss their advantages and limitations.
Inspired by evidence from the neural science, Riesenhuber \etal \cite{riesenhuber1996neural} explore neural networks with bottom-up analysis and top-down synthesis to explicitly address hierarchies in faces. 
Jin \etal \cite{jin2006context} design a probabilistic model for modeling part-whole hierarchies via perturbing a Markov backbone.
Capsule networks \cite{hinton2011transforming,sabour2017dynamic,hinton2018matrix} and their extensions \cite{lenssen2018group,kosiorek2019stacked,zhao20193d,sun2021canonical,sabour2021unsupervised} have been proposed to represent a part or an object in a hierarchy via a group of neurons called a \emph{capsule}. 
Given an image, a dynamic routing algorithm is exploited to determine the connections of capsules from consecutive levels.
Activated capsules at lower levels predict parameters of higher-level capsules.
A parse tree could then be created from the activated capsules and their connections.
However, the assumption that a capsule is permanently dedicated to a particular part or object restricts the expressiveness and flexibility of the model.
A conceptual framework for learning part-whole hierarchies, called \emph{GLOM}, is introduced in \cite{hinton2021represent}.
It consists of a large number of \emph{columns} which learn clustered representations at multiple levels and can be used to decode a part-whole hierarchy.
A few follow up works \cite{sun2021visual,abdurrafae2021hindsight} propose architectures like Transformers to implement this framework.
All these works focus on 2D images and some of them even aim for learning part-whole hierarchies in a unsupervised way, which is arguably too hard. 
Instead, we focus on learning hierarchies for 3D point clouds under a weakly supervised setup, which is more feasible.

There exists several approaches that learn part-whole hierarchies for 3D vision.
Mo \etal \cite{mo2019partnet} present a large-scale dataset of 3D objects analysis with annotated part or sub-part information.
Mo \etal \cite{mo2019structurenet} propose the StructureNet to encode 3D geometry via a hierarchy of n-ary graphs, where the hierarchical auto-encoder architecture captures both the geometry of parts and inter-part connections.
However, above works perform fully supervised learning which is less challenging and practical since it is hard to scale the dataset up.
Li \etal \cite{li2017grass} propose to represent 3D objects using a symmetry hierarchy and train a recursive auto-encoder network to predict its hierarchical structure.
Paschalidou \etal \cite{paschalidou2020learning} develop a model to jointly recover the geometry of a 3D object as a set of primitives and their latent hierarchical structures from a single view image.
Paschalidou \etal \cite{paschalidou2021neural} propose a 3D primitive representation learning model to parse 3D objects into semantically consistent part arrangements without part-level supervision.
These works perform purely unsupervised learning, \eg, learning to reconstruct 3D objects, and do not use point clouds to represent 3D objects, thus being significantly different from our method.

\subsection{Deep Learning on 3D Point Clouds}

Deep learning on 3D point clouds has bloomed in recent years, with various methods being proposed.
PointNet \cite{qi2017pointnet} and its extension PointNet++ \cite{qi2017pointnet++} are pioneer works that use multi-layer perceptrons (MLPs) to directly process points and achieve competitive results on point cloud classification and segmentation tasks.
PointCNN \cite{li2018pointcnn} generalizes CNNs by utilizing spatially local correlation within point clouds.
LDGCNN \cite{zhang2019linked} combines the hierarchical features at different layers in DGCNN to further improve the performance.
ClusterNet \cite{chen2019clusternet} extracts rotation-invariant features from $k$-nearest-neighbors (k-NN) of each point, and constructs hierarchical structures via the agglomerative hierarchical clustering.
ContinuousConv \cite{wang2018deep} and Relation-Shape CNN (RS-CNN) \cite{liu2019relation} extend regular-grid CNNs to the irregular setting.
PointConv \cite{wu2019pointconv} applies MLPs to local point coordinates to predict continuous weights and density functions of convolutional filters, which is permutation-invariant and translation-invariant.

On the other hand, graph neural networks (GNNs) \cite{scarselli2008graph} or graph convolution networks (GCNs) \cite{kipf2016semi} have been extensively applied to point clouds by treating points as nodes and constructing graphs such as k-NN graphs to learn point representations \cite{qi20173d,zhang2018graph,te2018rgcnn,wang2018local,wang2019dynamic,liu2019dynamic}.
Among them, Qi \etal \cite{qi20173d} first leverage message passing GNNs to learn point representations.
Zhang \etal \cite{zhang2018graph} propose a Graph-CNN model for point cloud classification, where localized graph convolutions with global and multi-resolution pooling are employed.
Te \etal \cite{te2018rgcnn} propose a regularized GCN architecture and design a smoothness prior of graph signals, which enforces graph-Laplacian-based smoothing in both the spectral and spatial domains.
Wang \etal \cite{wang2018local} propose a spectral graph convolution framework where a recursive clustering and pooling strategy is devised to aggregate information from clusters of nodes.
Wang \etal \cite{wang2019dynamic} adopt the EdgeConv operator and dynamically update node features and graphs during learning.
Although these methods do not explicitly model part-whole hierarchies, they still provide powerful backbones for representation learning on point clouds. 
In our work, we adopt PointNet as the backbone to obtain point representations.

Recently, several approaches have been proposed for semi-supervised point cloud learning \cite{wei2020multi,xu2020weakly,liu2021one,su2022weakly} and self-supervised/unsupervised point cloud learning \cite{gao2020graphter,rao2020global,sanghi2020info3d,xie2020pointcontrast,chen2021unsupervised,du2021self}, aiming to address the high cost of annotating point clouds for segmentation. 
Among them, semi-supervised approaches are more relevant to ours. 
Wei \etal \cite{wei2020multi} introduce a multi-path region mining module to generate pseudo point-level labels from a classification network trained with weak labels that indicate the classes that appeared in the input point cloud sample.
Xu \etal \cite{xu2020weakly} propose a segmentation approach which learns gradient approximation, exploits additional spatial-color smoothness constraints, and only requires a tiny fraction of points to be labeled for training.
Liu \etal \cite{liu2021one} introduce a relation network to generate the per-class prototype and explicitly model the similarities among nodes in order to generate pseudo labels for guiding the training.
Su \etal \cite{su2022weakly} propose a multi-prototype learning method, where each prototype serves as the classifier for one subclass of points.
All above methods only require some portion of labeled data.
In contrast, our method is orthogonal in that we learn part-whole hierarchies of point clouds with only the supervision of the top level.

\section{The Model}
\label{sec:model}
\begin{figure*}[t]
  \centering
  \includegraphics[width=\textwidth]{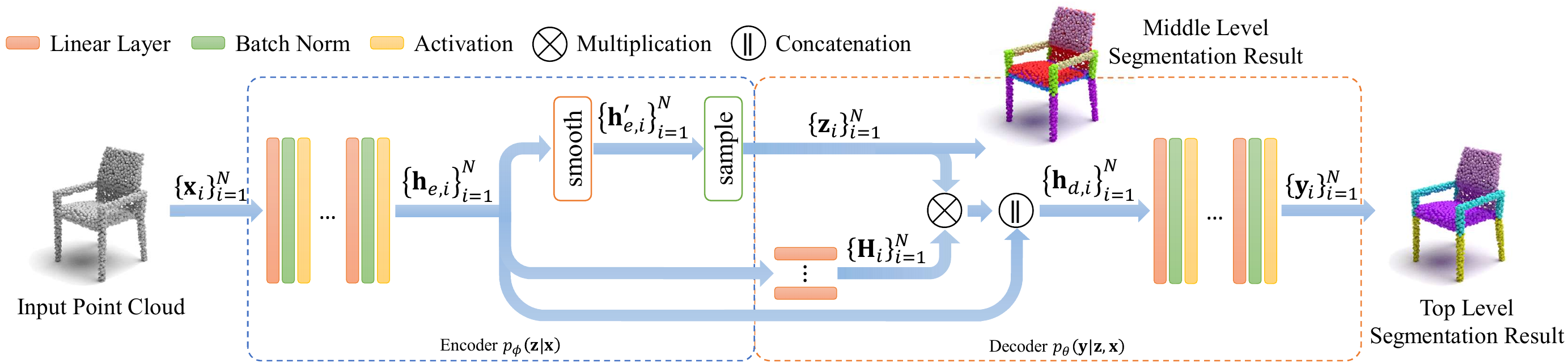}
  \caption{\textbf{The architecture of the proposed auto-encoder framework.} The probability $p_{\phi}(\z|\x)$ acts as the encoder, while $p_{\theta}(\y|\z,\x)$ plays the role of the label decoder, respectively. In our framework, the top level is supervised, while the middle level is unsupervised.}
  \label{fig:framework}
\end{figure*}

In this section, we first briefly introduce the common probabilistic model for classification in Section~\ref{subsec:the_common_model}, and then present the proposed latent variable model for learning part-whole hierarchies and classification in Section~\ref{subsec:the_model}. 

\subsection{Probabilistic Model for Classification}
\label{subsec:the_common_model}
We first introduce the common probabilistic model (without latent variables) for classification and then move on to the latent variable model.
In classical supervised learning of probabilistic classifiers, the typical learning objective is to minimize the following cross-entropy,
\begin{align}\label{eq:supervised_obj_CE}
& \min_{\theta} \quad \mathbb{E}_{q_{\text{data}}(\x)} \left[ H (q_{\text{data}}(\y\vert\x), p_{\theta}(\y\vert\x)) \right] \nonumber \\
\Leftrightarrow \quad & \max_{\theta} \quad \mathbb{E}_{q_{\text{data}}(\x)} \left[ - \mathbb{E}_{q_{\text{data}}(\y\vert\x)} \log  p_{\theta}(\y\vert\x) \right],
\end{align}
where $H(q, p)$ is the cross-entropy between two distributions $q$ and $p$.
$\x$ and $\y$ denote the input data and its label.
$p_{\theta}(\y\vert\x)$ is the classifier that needs to be learned and $\theta$ is the set of model parameters.
$q_{\text{data}}(\x)$ is the unknown data distribution.
$q_{\text{data}}(\y\vert\x)$ is the ground truth label distribution conditioned on data. 
The joint distribution $q_{\text{data}}(\y\vert\x)q_{\text{data}}(\x)$ specifies the generation process of observed data and labels.
Since we can only access data distribution via drawing i.i.d. samples and $q_{\text{data}}(\y\vert\x)$ is a Dirac delta measure, the objective in Eq. (\ref{eq:supervised_obj_CE}) is further approximated by the Monte Carlo (MC) estimation in practice,
\begin{align}
  \max_{\theta} \quad \frac{1}{N} \sum_{i=1}^{N} \log  p_{\theta}(\y = \y_i \vert \x = \x_i)
  \label{eq:supervised_obj},
\end{align}
where $\x_i \sim q_{\text{data}}(\x)$ is the $i$-th sampled training data and $\y_i$ is the corresponding ground truth class label.
For better readability, we will drop the sample index from now on. 
When $p_{\theta}(\y \vert \x)$ is parameterized by a deep neural network, the above probabilistic classifier achieves impressive results in many domains. 
However, it fails to capture complex or conceptual properties of the underlying data.
For example, in our context, a probabilistic classifier for segmenting 3D objects can not identify parts that are shared among them.

\subsection{Latent Variable Model for Part-Whole Hierarchies}\label{subsec:the_model}
To quantitatively model those unobservable factors that are important for both solving the downstream tasks (\eg, segmenting the whole 3D object) and describing the underlying data (\eg, segmenting the 3D object parts), we build a latent variable model.
In particular, we are given an input point cloud of a 3D object as $\x = \{\x_1, \x_2, \cdots, \x_m\}$, where $\x_i \in \mathbb{R}^3$ is a 3-dimension vector that specifies the 3D coordinates of the $i$-th point.
$m$ is the total number of points sampled from the underlying 3D object.
For 3D point cloud segmentation, given any point $\x_i$, we denote its semantic class label as a categorical random variable $\y_i \in \{1, 2, \cdots, K\}$.
To model part-whole hierarchies, we introduce latent variables $\z = \{\z_1, \z_2, \cdots, \z_m\}$ to represent the object part labels of individual points.
Here $\z_i \in \{1, 2, \cdots, C\}$ is again a categorical random variable, indicating the part label of the $i$-th point.
Since the part label is latent, we do not know its semantic meaning during training.
But after training, we can evaluate the accuracy of the learned part labels by grounding the latent variables to annotated part labels via bipartite matching.
More details about this matching step is explained in Section~\ref{subsubsec:benchmark}.
The total number of parts $C$ is a hyperparameter. 
Note that we only consider a two-level part-whole hierarchy due to the simplicity and the availability of annotated datasets.
We discuss how to generalize our method to more levels in Section~\ref{subsec:discuss}.

Since we introduce the latent variable, the conditional distribution of the classifier is decomposed as,
\begin{equation}\label{eq:latent_model}
  p(\y \vert\x) = \sum_{\z}p_{\theta}(\y\vert\z,\x) p_{\phi}(\z\vert\x),
\end{equation}
where the set of model parameters includes both $\theta$ and $\phi$.
We use these two separate notations for better explaining the learning algorithm in Section \ref{sec:learning}.
To ease the discussion, we call $p_{\phi}(\z\vert\x)$ as the \textit{encoder} which encodes the input point cloud $\x$ as latent codes (\ie, part labels) $\z$ in a probabilistic way.
For the tractability, we assume the conditional independence for part labels of individual points given the input,
\begin{equation}\label{eq:conditional_independence}
  p_{\phi}(\z\vert\x) = \prod_{i=1}^{m} p_{\phi}(\z_i\vert\x).
\end{equation}
Similarly, we call $p_{\theta}(\y\vert\z,\x)$ as the \textit{decoder} which decodes the semantic class label of individual points from the latent code $\z$ and the input $\x$.
For instance, if the sample $\x$ is a point cloud of a chair, we wish to encode the coordinates of individual points of $\x$ as the probabilities of being latent parts like ``surface" or ``frame".
Then from the latent codes and the original coordinates, we hope to decode the semantic label $\y$ as ``seat", ``back", or ``leg".
The overall model is illustrated in Fig.~\ref{fig:framework}.
This problem is an instance of the weakly supervised learning since we only have annotated segmentation labels $\y$ at the top level during training. 
We do not have any supervision on part label $\z$ in the bottom level.
The ground-truth annotations of part labels are only used for the final evaluation. 

\subsection{Encoder}
\label{subsec:encoder}

\begin{figure}
  \centering
  \includegraphics[width=\columnwidth]{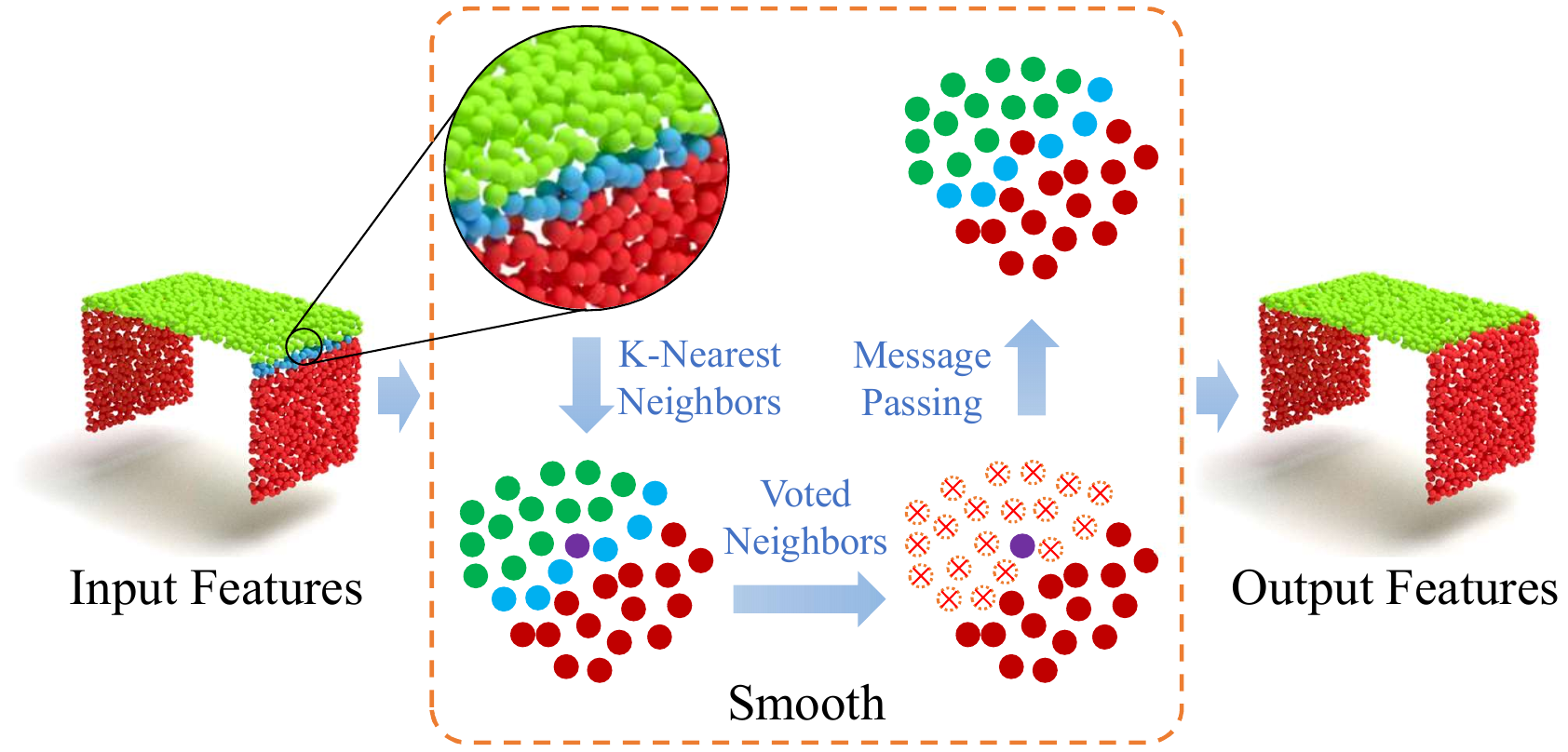}
  \caption{\textbf{An illustration of the smoothing module in the encoder.} We first construct the $K$-nearest-neighbor graph based on the input features, then use majority voting to obtain the voted neighborhood set, and finally employ a message passing layer to smooth the features.}
  \label{fig:gnn_smooth}
\end{figure}

We now introduce the specific design of our encoder which constructs the probability of the latent variable $\z_i$ given the input point $\x_i$.
As demonstrated in Fig.~\ref{fig:framework}, we first learn the feature representations $\h_{e,i}\in\mathbb{R}^{F_e}$ with $F_e$ channels of the input point $\x_i$ using a stack of linear layers, each followed by a batch normalization (BN) \cite{ioffe2015batch} and a ReLU activation \cite{nair2010rectified}.
The subscript $e$ emphasizes that these notations correspond to the encoder.
Based on these representations, we propose a message passing layer to propagate information among points so that the predicted latent part label of the current point depends on its local neighboring points, as illustrated in Fig.~\ref{fig:gnn_smooth}.
In particular, we treat each point $\x_i$ in the point cloud as a node and build a k-NN graph.
We denote $\mathcal{N}(i)$ as the set of $K$ nearest neighbors of node $i$.
Each point is also associated with a normal vector $\n_i \in \mathbb{R}^3$, which is estimated via plan fitting as discussed in detail in Section~\ref{subsubsec:datasets}.
We then perform the smoothing operation as
\begin{equation}
  \h_{e,i}^{'} = \W \left(\tilde{a}_{i,i}\h_{e,i} + \sum_{j\in\mathcal{N}(i) \setminus i}\tilde{a}_{i,j}\h_{e,j}\right),
\end{equation}
where $\W\in\mathbb{R}^{F_e \times F_e}$ is the learnable parameters, the normalized edge weight between node $i$ and node $j$ is denoted as $\tilde{a}_{i,j}=a_{i,j}/\sum_{k\in\mathcal{N}(i)}a_{i,k}$ where we use the absolute cosine similarity between normal vectors to compute the unnormalized edge weight, \ie, $a_{i,j}=|\cos(\n_i,\n_j)|$.
Specifically, we first collect the feature vectors $\h_j$ from the neighborhood $\mathcal{N}(i)$ and then perform a weighted sum based on their normal-vector-based similarities to the center node $i$.
Having obtained the smoothed feature representation $\h_{e,i}^{'}$, we employ one linear layer followed by a softmax to output the probability distribution $p_{\phi}(\z_i|\x_i)$.
After estimating the probability distribution $p_{\phi}(\z_i|\x_i)$, we perform sampling to acquire $L$ categorical (one-hot encoded) samples $\{\z_{i}^{(l)} \vert \z_{i}^{(l)} \in\mathbb{R}^{C}, l=1,\cdots,L \}$.
We use the superscript $(l)$ to denote the sample index.
As shown in Fig. \ref{fig:framework}, we treat the $L$ samples and the feature representations $\h_{e,i}$ as the input of the decoder to further estimate the label $\y_i$ for any point $\x_i$.

\subsection{Decoder}

At the decoder $p_{\theta}(\y|\z,\x)$, we first employ $C$ linear layers to map the representation $\h_{e,i}$ into $C$ separate representations in parallel which correspond to individual latent part classes.
We then concatenate these $C$ representations to obtain the feature representation matrix $\H_i \in\mathbb{R}^{F_d \times C}$ and compute the point representation $\h_{d,i}=\H_i \z_{i}$ that is conditioned on the latent variable $\z_i$.
Since $\z_i$ is a one-hot encoded categorical random variable, the conditioning is equivalent to selecting the corresponding column of $\H_i$ based on the latent part label.
It is clear that the representation $\h_{d,i}$ depends on the input point $\x_i$ and latent part label $\z_i$.

We then adopt several linear layers followed by a BN and a ReLU activation except that the last linear layer is followed by the softmax to estimate the probability distribution of label $\y_i$ according to Eq.~(\ref{eq:latent_model}).
\begin{equation}
\begin{split}
p(\y_i|\x_i) &\approx \frac{1}{L}\sum_{l=1}^{L}p_{\theta}(\y_i^{(l)}|\z_i^{(l)},\x_i) = \frac{1}{L}\sum_{l=1}^{L}p_{\theta}(\y_i^{(l)}|\h_{d,i}^{(l)}).
\end{split}
\end{equation}
We use $L$ samples to approximate the true classifier.

\section{Inference Algorithm}
\label{sec:inference}
In this section, we introduce the explored inference algorithms.
In the inference, given the input 3D point cloud $\x$, we would like to infer its latent lower-level part labels, \ie, computing $p_{\phi}(\z\vert\x)$, and infer its higher-level class label, \ie, computing $p(\y \vert \x)$.
Note that inferring the latent part labels is straightforward since we only need to run the forward pass of the encoder to obtain the probability $p_{\phi}(\z\vert\x)$ and then take the \emph{argmax} of the probability as the most probable latent part label.
However, the inference for the class label is tricky since it requires the marginalization over latent variable $\z$.
The marginalization is only tractable in low-dimension space.
In our case, $\z$ lies in the product space $\{1, 2, \cdots, C\}^{m}$ where $m$ is the number of points in the point cloud, thus making the inference intractable.
There exists a vast literature on approximated inference for probabilistic graphical models \cite{jordan1999introduction}.
For example, one can introduce a variational approximated distribution and perform variational inference by maximizing the evidence lower bound (ELBO).
However, this requires introducing either an iterative optimization procedure (standard variational inference \cite{hinton1993keeping,jordan1999introduction}) or a learned neural network based approximated variational distribution (the so-called amortized inference \cite{kingma2013auto,rezende2014stochastic}). 
Instead of the variational inference, we resort to the following two simple yet effective approximated inference algorithms.

\subsection{Most-Probable-Latent Inference}

Given model parameters $\theta$ and $\phi$, we could compute the conditional distribution of class labels via the following approximated inference.
We first perform the $\argmax$ operation on the encoder distribution $p_{\phi}(\z\vert\x)$ to acquire the latent variable $\z^{\ast}$ that indicates the most probable part label in the lower level.
This is why we name it as \emph{most-probable-latent inference} (MPL).
Then, we compute the probability as,
\begin{equation}
\begin{split}
 p(\y\vert\x) &\approx p_{\theta}(\y\vert\z^{\ast},\x) p_{\phi}(\z^{\ast}\vert\x) \\
 \quad \z^{\ast} &= \argmax_{\z} p_{\phi}(\z\vert\x).
 \label{eq:argmax_inference}
\end{split}
\end{equation}
We can predict the class label by taking the $\argmax$ of the $p(\y\vert\x)$ at the higher level.
It is easy to see that this algorithm provides a lower bound of the true conditional probability, 
\begin{equation}
\begin{split}
  p(\y\vert\x) &= \sum_{\z}p_{\theta}(\y\vert\z,\x) p_{\phi}(\z\vert\x) \\
  & \ge p_{\theta}(\y\vert\z^{\ast},\x) p_{\phi}(\z^{\ast}\vert\x).
\end{split}
\end{equation}
This approximated inference algorithm is deterministic and only requires one forward pass for the encoder and the decoder respectively, thus being very efficient.

\subsection{Monte Carlo Inference}

We can also approximately infer the label $\y$ based on the Monte Carlo estimation.
In particular, we have
\begin{equation}
\begin{split}
  p(\y\vert\x) &= \mathbb{E}_{p_{\phi}(\z\vert\x)} \left[ p_{\theta}(\y\vert\z,\x) \right] \approx \frac{1}{L}\sum_{l=1}^{L}p_{\theta}(\y \vert \z^{(l)},\x),  \\
  \z^{(l)} & \sim p_{\phi}(\z\vert\x). 
  \label{eq:monte_carlo_inference}
\end{split}
\end{equation}
In practice, we first compute the distribution $p_{\phi}(\z\vert\x)$ via one forward pass of the encoder.
Then, we sample $L$ part labels from it, compute the approximated $p_{\theta}(\y \vert \x)$ by running $L$ forward passes of the decoder, and finally take the $\argmax$.
The Monte Carlo estimator in Eq. (\ref{eq:monte_carlo_inference}) is unbiased.
Also, the law of large numbers guarantees that it is consistent.

\section{Learning Algorithm}
\label{sec:learning}
In this section, we delve into the details of learning algorithms.
To learn a probabilistic classifier, one typically maximizes the log likelihood of the target label $\y$ conditioned on the observed data $\x$ as shown in Eq.~(\ref{eq:supervised_obj}). 
In our case, based on Eq.~(\ref{eq:supervised_obj}) and Eq.~(\ref{eq:latent_model}), the resultant training objective is
\begin{align}
  \min_{\theta, \phi} \quad \mathcal{L}(\theta, \phi) \triangleq - \log \left( \sum_{\z}p_{\theta}(\y \vert \z,\x) p_{\phi}(\z \vert \x) \right)
  \label{eq:supervised_obj_LVM}.
\end{align}
Stochastic gradient descent (SGD) is often employed to learn the entire model end-to-end, which requires the gradient of the objective function in Eq.~(\ref{eq:supervised_obj_LVM}) \wrt parameters $\theta$ and $\phi$.
In the following, we introduce the gradient estimation methods under the previously proposed approximated inference methods, \ie, Eq.~(\ref{eq:argmax_inference}) or Eq.~(\ref{eq:monte_carlo_inference}), respectively. 

\subsection{Gradients for Most-Probable-Latent Inference}
In particular, the objective function in this case is,
\begin{equation}
\begin{split}
  \mathcal{L}(\theta, \phi) &\approx - \log \left( p_{\theta}(\y \vert \z^{\ast},\x) p_{\phi}(\z^{\ast} \vert \x) \right) \\
  \z^{\ast} &= \argmax_{\z} p_{\phi}(\z\vert\x).
  \label{eq:supervised_obj_LVM_deterministic}
\end{split}
\end{equation}
Let us first assume that $\z^{\ast}$ is differentiable \wrt $\phi$.
Then we have the following gradients,
\begin{equation}
\begin{split}
  \frac{\partial\mathcal{L}(\theta,\phi)}{\partial\theta} &= - \frac{\partial \log p_{\theta}(\y \vert \z^{\ast},\x)}{\partial \theta} \\
  \frac{\partial\mathcal{L}(\theta,\phi)}{\partial\phi} &= - \frac{\partial \log p_{\theta}(\y \vert \z^{\ast},\x)}{\partial \z^{\ast}} \frac{\partial \z^{\ast}}{\partial \phi} - \frac{\partial \log p_{\phi}(\z^{\ast} \vert \x)}{\partial \phi}.
  \label{eq:gradient_most_probable_latent}
\end{split}
\end{equation}
If the previous assumption holds, then we can use Eq.~(\ref{eq:gradient_most_probable_latent}) to compute the SGD update.
However, $\z$ in our case is a discrete (or categorical) random variable, thus breaking the assumption that $\z^{\ast}$ is differentiable \wrt $\phi$.

\subsubsection{Straight-Through Estimator}
To overcome the above challenge, we resort to the straight-through estimator (STE) \cite{bengio2013estimating}.
STE is a simple and widely used technique for estimating gradients of discrete random variables.
The core idea is to relax the discrete random variable to the continuous domain and ignore the non-differentiable transformation function when computing the gradient.
In the most-probable-latent inference, given the output probability distribution of the encoder network $p_{\phi}(\z \vert \x) = \bm{\pi} = [\bm{\pi}_1,...,\bm{\pi}_C]$ (typically obtained by applying the softmax), the latent part label $\z^{\ast}$ is computed as $\z^{\ast} = \argmax_{i} \bm{\pi}_i$.
STE just approximates the gradient by ignoring the $\argmax$ in the backward pass, 
\begin{equation}
\begin{split}
    \frac{\partial\mathcal{L}(\theta,\phi)}{\partial\phi} &= - \frac{\partial \log p_{\theta}(\y \vert \z^{\ast},\x)}{\partial \z^{\ast}} \frac{\partial \z^{\ast}}{\partial \phi} - \frac{\partial \log p_{\phi}(\z^{\ast} \vert \x)}{\partial \phi} \\
    &\approx - \frac{\partial \log p_{\theta}(\y \vert \z^{\ast},\x)}{\partial \z^{\ast}} \frac{\partial \bm{\pi}_{\z^{\ast}}}{\partial \phi} - \frac{\partial \log p_{\phi}(\z^{\ast} \vert \x)}{\partial \phi},
  \label{eq:gradient_ste}    
\end{split}
\end{equation}
where $\bm{\pi}_{\z^{\ast}} = \max_{i} \bm{\pi}_{i}$.
Albeit being simple, there are no theoretical guarantees for STE on, \eg, consistency and unbiasedness.
But it still serves as a popular baseline method due to its good empirical performances on some tasks.

\subsection{Gradients for Monte Carlo Inference}
For Monte Carlo inference in Eq.~(\ref{eq:monte_carlo_inference}), we can derive the gradient \wrt decoder parameters $\theta$ as follows,
\begin{equation}
\begin{split}
  \frac{\partial\mathcal{L}(\theta,\phi)}{\partial\theta}&=-\frac{\partial\log\sum_{\z}p_{\theta}(\y\vert\z,\x)p_{\phi}(\z\vert\x)}{\partial\theta} \\
  &= - \frac{1}{Z} \sum_{\z}\frac{\partial p_{\theta}(\y\vert\z,\x)}{\partial\theta}p_{\phi}(\z\vert\x) \\
  &= -\frac{1}{Z} \mathbb{E}_{p_{\phi}(\z\vert\x)} \left[ \frac{\partial p_{\theta}(\y\vert\z,\x)}{\partial\theta} \right],
  \label{eq:gradient_decoder}
\end{split}
\end{equation}
where $Z=\sum_{\z}p_{\theta}(\y\vert\z,\x)p_{\phi}(\z\vert\x) = p(\y\vert\x)$.
Since the decoder $p_{\theta}(\y\vert\z,\x)$ will be constructed using a differentiable neural network, the gradient term within the expectation is available.
Therefore, one can estimate the gradient \wrt decoder parameters $\theta$ as,
\begin{equation}\label{eq:gradient_decoder_MC}
  \frac{\partial\mathcal{L}(\theta,\phi)}{\partial\theta} \approx -\frac{1}{\hat{Z}L} \sum_{l=1}^{L}  \frac{\partial p_{\theta}(\y\vert\z^{(l)},\x)}{\partial\theta},
\end{equation}
where $\hat{Z} = \frac{1}{L} \sum_{l=1}^{L} p_{\theta}(\y \vert \z^{(l)},\x)$ and $\z^{(l)} \sim p_{\phi}(\z\vert\x)$. 
Now let us turn to estimating the gradients w.r.t. the encoder parameters $\phi$,
\begin{equation}
\begin{split}
  \frac{\partial\mathcal{L}(\theta,\phi)}{\partial\phi}&= -\frac{\partial\log\sum_{\z}p_{\theta}(\y\vert\z,\x)p_{\phi}(\z\vert\x)}{\partial\phi} \\
  &=- \frac{1}{Z} \sum_{\z}p_{\theta}(\y\vert\z,\x)\frac{\partial p_{\phi}(\z\vert\x)}{\partial\phi}.
  \label{eq:gradient_encoder_MC}
\end{split}
\end{equation}
Since we use a differentiable neural network to construct the encoder, the gradient ${\partial p_{\phi}(\z\vert\x)}/{\partial\phi}$ is available.
However, comparing Eq.~(\ref{eq:gradient_decoder}) to Eq.~(\ref{eq:gradient_encoder_MC}), we can find that there does not exist an expectation term for the encoder, which prevents us from using Monte Carlo estimation as in Eq.~(\ref{eq:gradient_decoder_MC}).
To address this issue, we explore the REINFORCE and pathwise gradient estimators.

\subsubsection{REINFORCE Gradient Estimator}

REINFORCE estimator \cite{williams1992simple}, \aka the score function estimator \cite{rubinstein1996score} or the likelihood ratio method \cite{glynn1990likelihood}, is a widespread and general-purpose method to deal with the aforementioned estimation problem. 
It leverages a simple log-derivative trick and applies to both discrete and continuous random variables.

In our context, when the probability mass function $p_{\phi}(\z\vert\x)$ is differentiable w.r.t. encoder parameters $\phi$, the log-derivative trick is,
\begin{equation}
  \frac{\partial p_{\phi}(\z\vert\x)}{\partial\phi}=p_{\phi}(\z\vert\x)\frac{\partial\log p_{\phi}(\z\vert\x)}{\partial\phi}.
\end{equation}
The gradients of the encoder in Eq.~(\ref{eq:gradient_encoder_MC}) thus become,
\begin{equation}
\begin{split}
  \frac{\partial\mathcal{L}(\theta,\phi)}{\partial\phi} &= - \frac{1}{Z} \sum_{\z}p_{\theta}(\y\vert\z,\x)p_{\phi}(\z\vert\x)\frac{\partial\log p_{\phi}(\z\vert\x)}{\partial\phi} \\
  &= - \frac{1}{Z} \mathbb{E}_{p_{\phi(\z\vert\x)}}\left[\frac{\partial\log p_{\phi}(\z\vert\x)}{\partial\phi}p_{\theta}(\y\vert\z,\x)\right].
\label{eq:REINFORCE_grad}
\end{split}
\end{equation}
Now we have the gradient rewritten in the form of an expectation, which enables the usage of Monte Carlo estimation.
Note that REINFORCE estimator in our context has a close relationship with reinforcement learning (RL) as below,
\begin{equation}
  -\frac{\partial\mathcal{L}(\theta,\phi)}{\partial\phi} = \mathbb{E}_{p_{\phi(\z\vert\x)}}\left[\underbrace{\frac{p_{\theta}(\y\vert\z,\x)}{Z}}_{\text{Reward}} \frac{\partial }{\partial \phi} \log \underbrace{p_{\phi}(\z\vert\x)}_{\text{Policy}} \right].
\end{equation}
The discrete part label $\z$ can be treated as an action, the encoder $p_{\phi}(\z\vert\x)$ thus corresponds to the policy network, and the term involving decoder $p_{\theta}(\y\vert\z,\x) / Z$ corresponds to the reward function.
Our goal is to learn the policy parameters $\phi$ to maximize our expected reward.

We can now construct a Monte Carlo estimator as below,
\begin{equation}\label{eq:REINFORCE_MC}
  \frac{\partial\mathcal{L}(\theta,\phi)}{\partial\phi} \approx -\frac{1}{\hat{Z}L}\sum_{i=1}^{L}\frac{\partial\log p_{\phi}(\z^{(l)} \vert \x)}{\partial\phi}p_{\theta}(\y \vert\z^{(l)},\x),
\end{equation}
where $\hat{Z} = \frac{1}{L} \sum_{l=1}^{L} p_{\theta}(\y \vert \z^{(l)},\x)$ and $\z^{(l)} \sim p_{\phi}(\z\vert\x)$.

\textbf{Variance Reduction} The REINFORCE estimator is unbiased but may come with high variances.
One can further reduce the variance via the control variate method \cite{mcbook}, \aka baseline in RL.
In particular, if we introduce another random variable $B$ (\ie, the so called \emph{control variate}) which does not depend on $\z$, we have the following identity,
\begin{equation}
  \mathbb{E}_{p_{\phi(\z\vert\x)}}\left[ \frac{\partial\log p_{\phi}(\z\vert\x)}{\partial\phi} B \right] = 0.
\end{equation}
Relying on the above observation, we can derive an alternative REINFORCE estimator:
\begin{equation}
 \frac{\partial\mathcal{L}(\theta,\phi)}{\partial\phi}\approx \frac{1}{\hat{Z} L}\sum_{l=1}^{L}\left[\frac{\partial\log p_{\phi}(\z^{(l)} \vert \x)}{\partial\phi}\left(p_{\theta}(\y \vert \z^{(l)},\x) - B\right)\right].
\end{equation}
In practice, we found setting $B=1$ works well.

\subsubsection{Pathwise Gradient Estimator}

The pathwise gradient estimator \cite{schulman2015gradient}, \aka reparameterization trick \cite{kingma2013auto}, leverages the change-of-varaible technique to rewrite the target random variable as a function of a base random variable which is typically drawn from a simple distribution. 
In our context, since $\z$ is a discrete (or categorical) random variable indicating the part label, one can reparameterize the sample path using the following Gumbel-max trick~\cite{papandreou2011perturb}.
\begin{equation}
\begin{split}
  \z &= g_{\phi}(\x, \bm{\epsilon}) = \argmax \left( \log(\bm{\pi}) + \bm{\epsilon}) \right) \\
  \bm{\epsilon}[i] &\sim \mathrm{Gumbel}(0, 1) \qquad \forall i = 1, \cdots, C,
\end{split}    
\end{equation}
where $\bm{\pi}=p_{\phi}(\z \vert \x)=[\bm{\pi}_1,...,\bm{\pi}_C]$ is again the probability output by the encoder via the softmax, thus depending on $\x$ and $\phi$ in a differentiable way.
The base distribution of $\bm{\epsilon}=[\bm{\epsilon}_1,...,\bm{\epsilon}_C]$ is the standard Gumbel distribution.
$\bm{\epsilon}[i]$ denotes the $i$-th element of the vector $\bm{\epsilon}$.
However, the corresponding function $g_{\phi}(\x, \bm{\epsilon})$ is not differentiable \wrt $\phi$ due to the argmax operator.
To facilitate gradient-based learning, the Gumbel-softmax trick \cite{jang2016categorical,maddison2016concrete} instead approximates the argmax operator using a softmax operator which preserves the differentiability.
Specifically, with the one-hot encoding of $\z$, we have
\begin{equation}
\begin{split}\label{eq:gumbel_softmax}
  \z[i] &\approx g_{\phi}(\x, \bm{\epsilon})[i] =\frac{\exp\left\{(\log(\bm{\pi}[i]) + \bm{\epsilon}[i])/\tau\right\}}{\sum_{j=1}^{C}\exp\left\{(\log(\bm{\pi}[j]) + \bm{\epsilon}[j])/\tau\right\}},
\end{split}    
\end{equation}
where the base distribution is again the standard Gumbel distribution, \ie, $\forall i$, $\bm{\epsilon}[i] \sim \mathrm{Gumbel}(0, 1)$, and $\tau$ is the temperature parameter.
We perform the ablation study on the temperature in Section~\ref{subsubsec:tau_ablation}.
$\z[i] = 1$ indicates that the part label is $i$.
In Gumbel-max trick, $\z$ corresponds to the vertices of the probability simplex, thus taking discrete values from a finite set.
In Gumbel-softmax trick, $\z$ is relaxed to a continuous domain, \ie, the whole probability simplex.

Based on the Gumbel-softmax reparameterization in Eq. (\ref{eq:gumbel_softmax}), we can estimate the gradient under Monte Carlo inference.
Specifically, we have 
\begin{equation}
\begin{split}
  \frac{\partial\mathcal{L}(\theta,\phi)}{\partial\phi} &\approx -\frac{\partial\log \left( \frac{1}{L}\sum_{l=1}^{L}p_{\theta}(\y \vert \z^{(l)},\x) \right)}{\partial\phi} \\
    &= - \frac{1}{L\hat{Z}} \sum_{l=1}^{L} \frac{\partial p_{\theta}(\y \vert \z^{(l)},\x)}{\partial \z^{(l)}} \frac{\partial \z^{(l)}}{\partial \phi},
  \label{eq:gradient_encoder_path_wise}
\end{split}
\end{equation}
where $\hat{Z} = \frac{1}{L} \sum_{l=1}^{L} p_{\theta}(\y \vert \z^{(l)},\x)$ and $\z^{(l)}$ is drawn via the Gumbel-softmax distribution.
Note that the pathwise gradient estimator is biased due to the fact the Gumbel-softmax trick uses softmax to approximate the argmax operator.
But the variance of the estimator is often found to be low.

\section{Experiments}
\label{sec:experiments}
\begin{table*}[t]
  \centering
  \caption{Multi-level point cloud segmentation results in terms of accuracy (\%). We denote the best performance in \textbf{bold} and second best with \underline{underline}.}
  \label{tab:seg_results}
  \begin{tabularx}{\textwidth}{lYYYYYYYY}
  \hline
  \multicolumn{9}{c}{\textbf{Middle Level}} \\ \hline
  \multicolumn{1}{c|}{\textbf{Methods}} & \textbf{Bed} & \textbf{Chair} & \textbf{Clock} & \textbf{Dishwasher} & \textbf{Display} & \textbf{Door} & \textbf{Earphone} & \textbf{Faucet} \\ \hline
  \multicolumn{1}{c|}{\textbf{\#Samples}} & 36 & 1217 & 96 & 51 & 191 & 51 & 53 & 131 \\
  \multicolumn{1}{c|}{\textbf{\#Classes}} & 10 & 30 & 11 & 5 & 4 & 4 & 10 & 12 \\ \hline
  \multicolumn{1}{r|}{\textbf{PointNet} \cite{qi2017pointnet}} & 42.23 & 48.97 & 51.67 & 63.10 & 83.88 & 79.86 & 58.37 & 56.22 \\
  \multicolumn{1}{r|}{\textbf{PointNet++} \cite{qi2017pointnet++}} & 36.97 & 45.68 & 43.51 & 52.31 & 65.09 & 80.06 & 55.13 & 51.62 \\
  \multicolumn{1}{r|}{\textbf{PointCNN} \cite{li2018pointcnn}} & 34.89 & 38.39 & 40.96 & 45.95 & 67.71 & 73.60 & 43.53 & 42.84 \\
  \multicolumn{1}{r|}{\textbf{PointConv} \cite{wu2019pointconv}} & 38.04 & 43.95 & 51.60 & 59.70 & 82.52 & 80.46 & 49.36 & 50.82 \\
  \cdashline{1-9}[2pt/2pt]
  \multicolumn{1}{c|}{\textbf{MC-Pathwise}} & 43.13 & \underline{68.76} & \textbf{58.81} & 65.08 & 84.10 & \underline{82.50} & \underline{58.45} & 53.94 \\
  \multicolumn{1}{c|}{\textbf{MPL-STE}} & \textbf{45.99} & 67.64 & 58.26 & \textbf{66.40} & \underline{85.39} & \textbf{83.73} & \textbf{58.52} & \textbf{59.04} \\
  \multicolumn{1}{c|}{\textbf{MC-REINFORCE}} & \underline{43.79} & \textbf{73.96} & \underline{58.58} & \underline{65.15} & \textbf{87.93} & 81.27 & 56.75 & \underline{56.31} \\ \hline
  \multicolumn{1}{c|}{\textbf{Methods}} & \textbf{Knife} & \textbf{Lamp} & \textbf{Microwave} & \textbf{Refrigerator} & \textbf{Storage} & \textbf{Table} & \textbf{Trash Can} & \textbf{Vase} \\ \hline
  \multicolumn{1}{c|}{\textbf{\#Samples}} & 77 & 397 & 39 & 31 & 450 & 1661 & 63 & 233 \\
  \multicolumn{1}{c|}{\textbf{\#Classes}} & 10 & 28 & 5 & 6 & 19 & 42 & 11 & 6 \\ \hline
  \multicolumn{1}{r|}{\textbf{PointNet} \cite{qi2017pointnet}} & 66.96 & 71.28 & 63.77 & 54.08 & 48.50 & 45.07 & 52.39 & 66.16 \\
  \multicolumn{1}{r|}{\textbf{PointNet++} \cite{qi2017pointnet++}} & 69.11 & 67.22 & 50.59 & 49.44 & 31.78 & 37.21 & 43.48 & 62.00 \\
  \multicolumn{1}{r|}{\textbf{PointCNN} \cite{li2018pointcnn}} & 46.29 & 49.13 & 53.45 & 47.67 & 37.63 & 41.85 & 31.38 & 56.90 \\
  \multicolumn{1}{r|}{\textbf{PointConv} \cite{wu2019pointconv}} & 54.82 & 66.60 & 64.65 & 58.32 & 39.43 & 37.58 & 49.78 & 75.09 \\
  \cdashline{1-9}[2pt/2pt]
  \multicolumn{1}{c|}{\textbf{MC-Pathwise}} & 69.61 & \underline{73.61} & 64.83 & \underline{59.54} & \underline{54.64} & \underline{70.15} & 55.87 & 75.82 \\
  \multicolumn{1}{c|}{\textbf{MPL-STE}} & \textbf{73.13} & 69.99 & \underline{65.65} & 59.38 & \textbf{57.72} & 69.50 & \textbf{64.30} & \underline{77.91} \\
  \multicolumn{1}{c|}{\textbf{MC-REINFORCE}} & \underline{70.87} & \textbf{76.51} & \textbf{65.72} & \textbf{59.83} & 54.63 & \textbf{71.92} & \underline{63.53} & \textbf{80.25} \\ \hline \hline
  \multicolumn{9}{c}{\textbf{Top Level}} \\ \hline
  \multicolumn{1}{c|}{\textbf{Methods}} & \textbf{Bed} & \textbf{Chair} & \textbf{Clock} & \textbf{Dishwasher} & \textbf{Display} & \textbf{Door} & \textbf{Earphone} & \textbf{Faucet} \\ \hline
  \multicolumn{1}{c|}{\textbf{\#Samples}} & 36 & 1217 & 96 & 51 & 191 & 51 & 53 & 131 \\
  \multicolumn{1}{c|}{\textbf{\#Classes}} & 4 & 7 & 6 & 3 & 3 & 3 & 6 & 8 \\ \hline
  \multicolumn{1}{r|}{\textbf{PointNet} \cite{qi2017pointnet}} & 72.57 & 89.91 & 80.94 & 94.61 & 94.66 & 82.41 & 85.40 & 82.46 \\
  \multicolumn{1}{r|}{\textbf{PointNet++} \cite{qi2017pointnet++}} & 75.36 & 90.32 & 82.18 & 93.34 & 93.64 & 80.88 & 84.68 & 81.89 \\
  \multicolumn{1}{r|}{\textbf{PointCNN} \cite{li2018pointcnn}} & 76.01 & \textbf{92.01} & 75.41 & 93.59 & \underline{96.16} & 79.69 & 81.20 & 81.82 \\
  \multicolumn{1}{r|}{\textbf{PointConv} \cite{wu2019pointconv}} & 73.36 & 91.42 & 82.01 & 87.16 & \textbf{96.33} & 80.22 & 79.29 & 82.26 \\
  \cdashline{1-9}[2pt/2pt]
  \multicolumn{1}{c|}{\textbf{MC-Pathwise}} & 78.83 & 91.15 & \textbf{83.63} & 95.19 & 95.15 & \underline{85.59} & \underline{86.11} & 83.53 \\
  \multicolumn{1}{c|}{\textbf{MPL-STE}} & \underline{79.19} & 91.21 & \underline{82.85} & \underline{95.44} & 94.22 & \textbf{86.53} & \textbf{86.77} & \textbf{84.79} \\
  \multicolumn{1}{c|}{\textbf{MC-REINFORCE}} & \textbf{79.72} & \underline{91.81} & 82.33 & \textbf{95.48} & 95.96 & 83.99 & 85.78 & \underline{84.28} \\ \hline
  \multicolumn{1}{c|}{\textbf{Methods}} & \textbf{Knife} & \textbf{Lamp} & \textbf{Microwave} & \textbf{Refrigerator} & \textbf{Storage} & \textbf{Table} & \textbf{Trash Can} & \textbf{Vase} \\ \hline
  \multicolumn{1}{c|}{\textbf{\#Samples}} & 77 & 397 & 39 & 31 & 450 & 1661 & 63 & 233 \\
  \multicolumn{1}{c|}{\textbf{\#Classes}} & 5 & 18 & 3 & 3 & 7 & 12 & 5 & 4 \\ \hline
  \multicolumn{1}{r|}{\textbf{PointNet} \cite{qi2017pointnet}} & 80.12 & 75.54 & 93.76 & 92.29 & 78.96 & 90.18 & 80.38 & 88.90 \\
  \multicolumn{1}{r|}{\textbf{PointNet++} \cite{qi2017pointnet++}} & \textbf{82.64} & 76.20 & 90.53 & 88.85 & 81.03 & 90.53 & 80.48 & 88.64 \\
  \multicolumn{1}{r|}{\textbf{PointCNN} \cite{li2018pointcnn}} & 64.41 & 63.34 & 92.83 & 87.72 & \textbf{84.45} & 91.05 & 73.08 & 88.81 \\
  \multicolumn{1}{r|}{\textbf{PointConv} \cite{wu2019pointconv}} & 69.38 & 75.12 & 97.23 & 89.98 & \underline{82.22} & \underline{91.62} & 79.70 & \textbf{91.98} \\
  \cdashline{1-9}[2pt/2pt]
  \multicolumn{1}{c|}{\textbf{MC-Pathwise}} & \underline{82.36} & 76.53 & 97.39 & \underline{93.15} & 80.89 & 91.61 & \underline{85.38} & 88.06 \\
  \multicolumn{1}{c|}{\textbf{MPL-STE}} & 79.85 & \underline{76.71} & \underline{97.50} & 91.21 & 81.16 & 91.18 & \textbf{86.92} & \underline{88.96} \\
  \multicolumn{1}{c|}{\textbf{MC-REINFORCE}} & 80.29 & \textbf{78.58} & \textbf{97.91} & \textbf{95.60} & 81.07 & \textbf{92.39} & 85.22 & 87.85 \\ \hline
  \end{tabularx}
\end{table*}

\begin{table}[t]
\centering
\caption{Multi-level point cloud segmentation results in terms of mean accuracy (\%) across categories.}
\label{tab:mean_seg_acc}
\begin{tabular}{c|cc}
\hline
  & \textbf{Middle Level} & \textbf{Top Level} \\ \hline
PointNet \cite{qi2017pointnet} & 59.53 & 85.19 \\
PointNet++ \cite{qi2017pointnet++} & 52.58 & 85.07 \\
PointCNN \cite{li2018pointcnn} & 47.01 & 82.60 \\
PointConv \cite{wu2019pointconv} & 56.42 & 84.33 \\ \hline
MC-Pathwise & 64.93 & 87.16 \\
MPL-STE & 66.41 & 87.16 \\
MC-REINFORCE & \textbf{66.69} & \textbf{87.39} \\ \hline
\end{tabular}
\end{table}

In this section, we evaluate the proposed model on weakly-supervised multi-level 3D point cloud segmentation task and perform ablation study on our design choices.

\subsection{Multi-Level Point Cloud Segmentation}

\subsubsection{Datasets}
\label{subsubsec:datasets}

We use the PartNet dataset \cite{mo2019partnet} to evaluate our model.
This dataset consists of $573,585$ fine-grained part annotations for $26,671$ shapes across $24$ different 3D object categories.
In this dataset, most of the 3D objects are annotated at three levels: coarse, middle, and fine-grained levels.
Among the $24$ object categories, all of them have the coarse-level segmentation labels, while $9$ have the middle-level and $17$ have the fine-level ones.

We choose $16$ categories and the corresponding coarse- and fine-grained level as the middle and top levels for experiments, including \texttt{Bed}, \texttt{Chair}, \texttt{Clock}, \texttt{Dishwasher}, \texttt{Display}, \texttt{Door}, \texttt{Earphone}, \texttt{Faucet}, \texttt{Knife}, \texttt{Lamp}, \texttt{Microwave}, \texttt{Refrigerator}, \texttt{Storage}, \texttt{Table}, \texttt{Trash Can}, and \texttt{Vase}.
For each model, $2,048$ points are sampled from the original object.
Following the previous literature, we train separate networks for each object category to better handle the imbalance issue across categories. 
We follow the official train/test split as recommended by the authors to conduct the experiments.
Note that we do not utilize any data augmentation mechanisms to train the network.

We also estimate the normal vector $\n_i$ per point for latent variable encoder, as described in  Section~\ref{subsec:encoder}.
Specifically, given any point $\x_i\in\mathbb{R}^3$ and its $K$ nearest neighbors $\mathcal{N}(i)$, we fit a plane via the least square method.
The normal vector $\n_i\in\mathbb{R}^{3}$ of the fitted plane is thus treated as the normal vector of the point $\x_i$.

\subsubsection{Evaluation Metrics}
\label{subsubsec:metrics}

We treat the 3D point cloud segmentation as the point-wise classification problem. 
We report the overall accuracy (OA), \ie, the mean accuracy for all test points in the dataset, 
for the performance evaluation.
Since the middle level is unsupervised, OA cannot be directly computed.
Therefore, we first estimate the latent part label of every point in each 3D point cloud through the network at the middle level, then use the Hungarian algorithm \cite{kuhn1955hungarian} to match the predicted part label with the ground-truth, and finally calculate the OA between the matched labels and the ground-truth ones.
Note that the matching here is done in an instance-wise fashion, \ie, running the Hungarian algorithm for each point cloud separately.

\begin{figure*}[htbp]
  \centering
  \subfigure[Bed]{\includegraphics[width=0.33\textwidth]{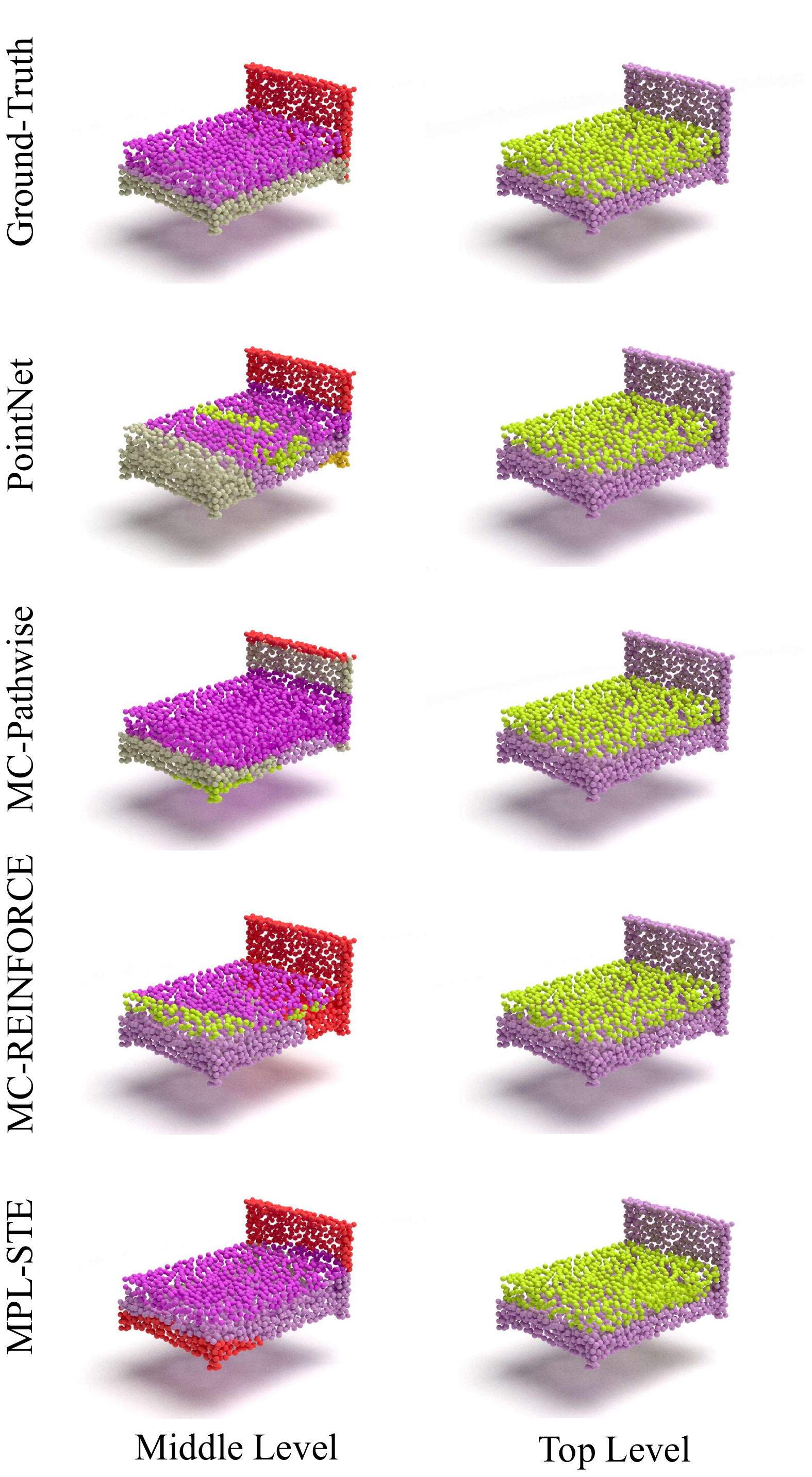}}
  \subfigure[Chair]{\includegraphics[width=0.33\textwidth]{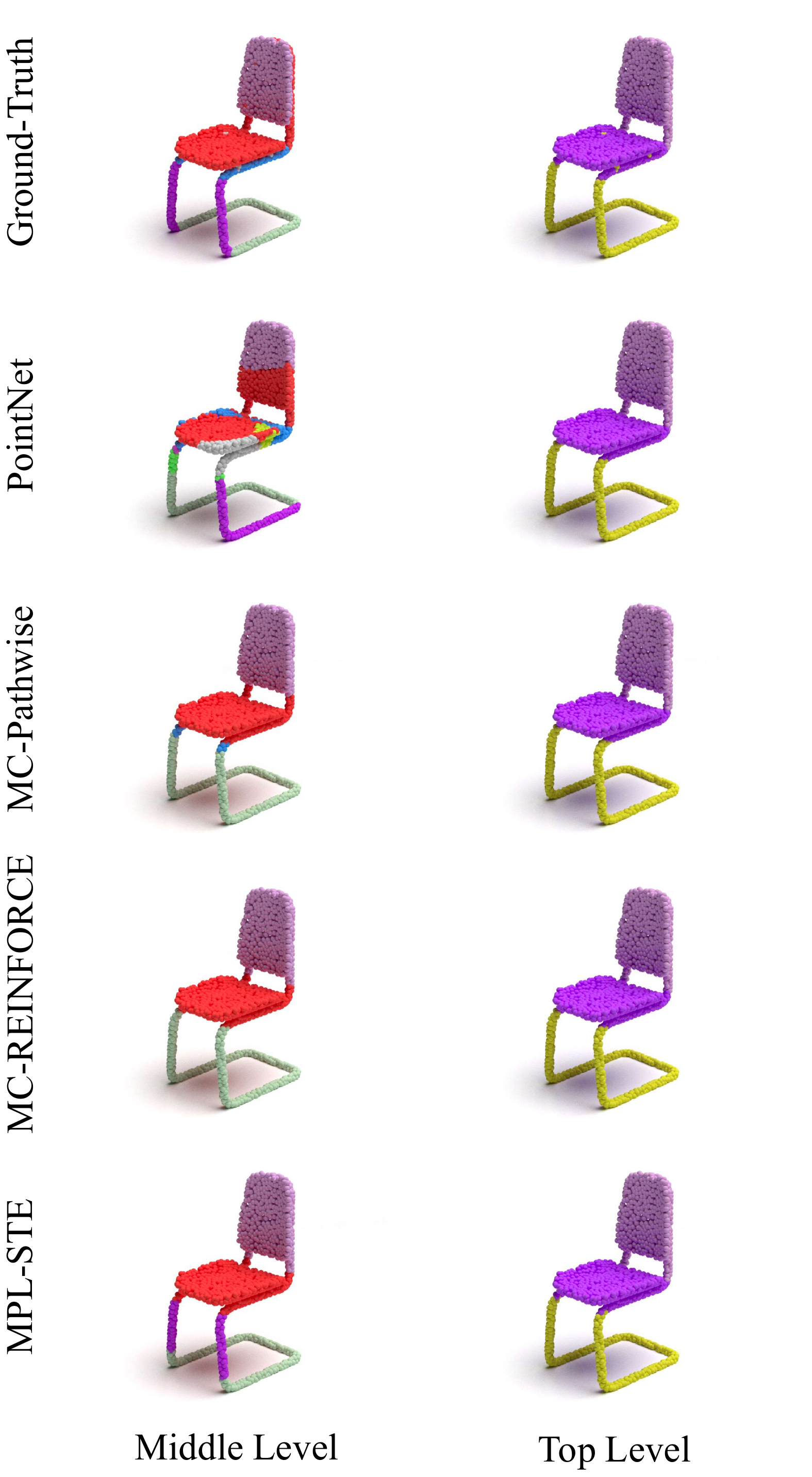}}
  \subfigure[Chair]{\includegraphics[width=0.33\textwidth]{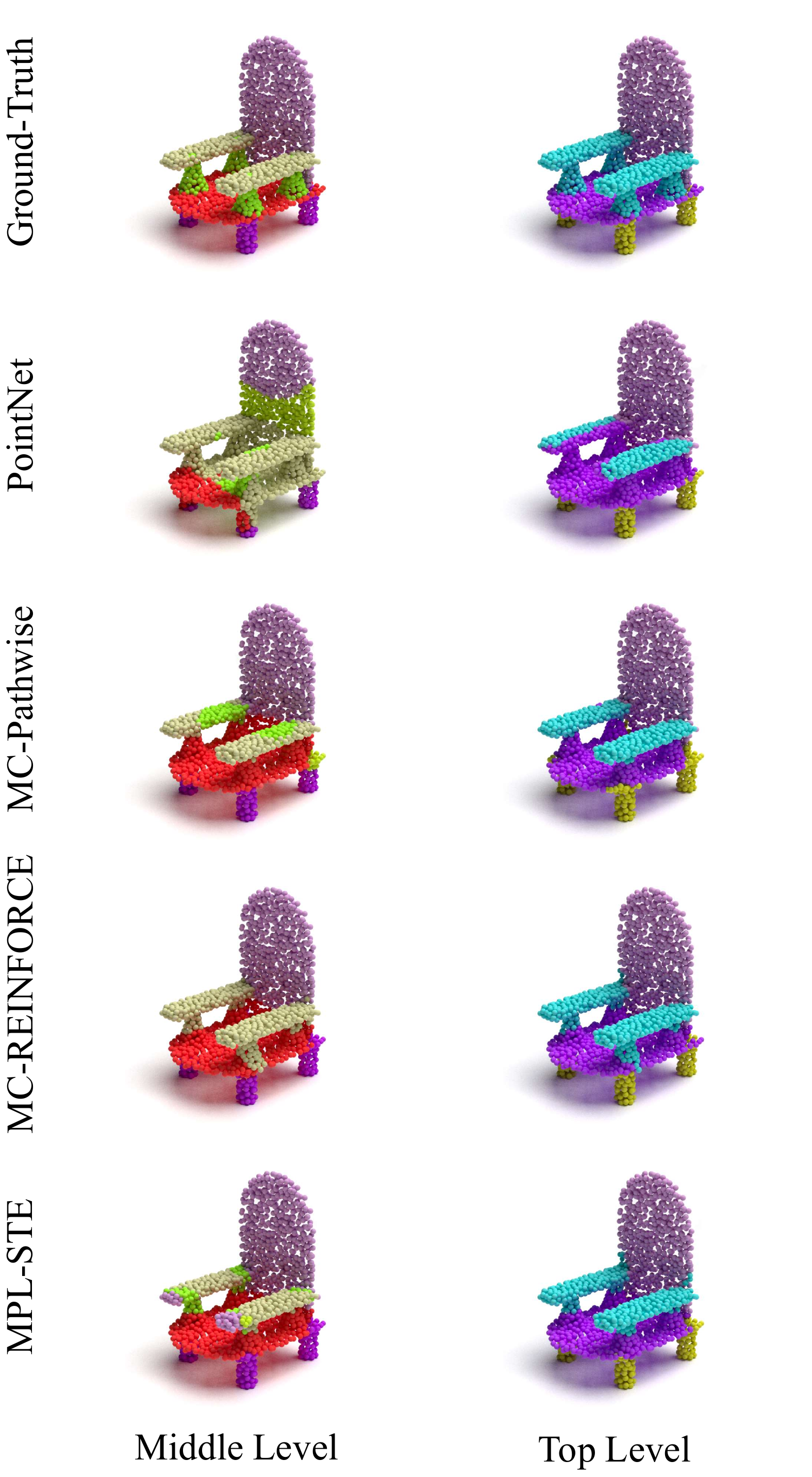}}
  \subfigure[Dishwasher]{\includegraphics[width=0.33\textwidth]{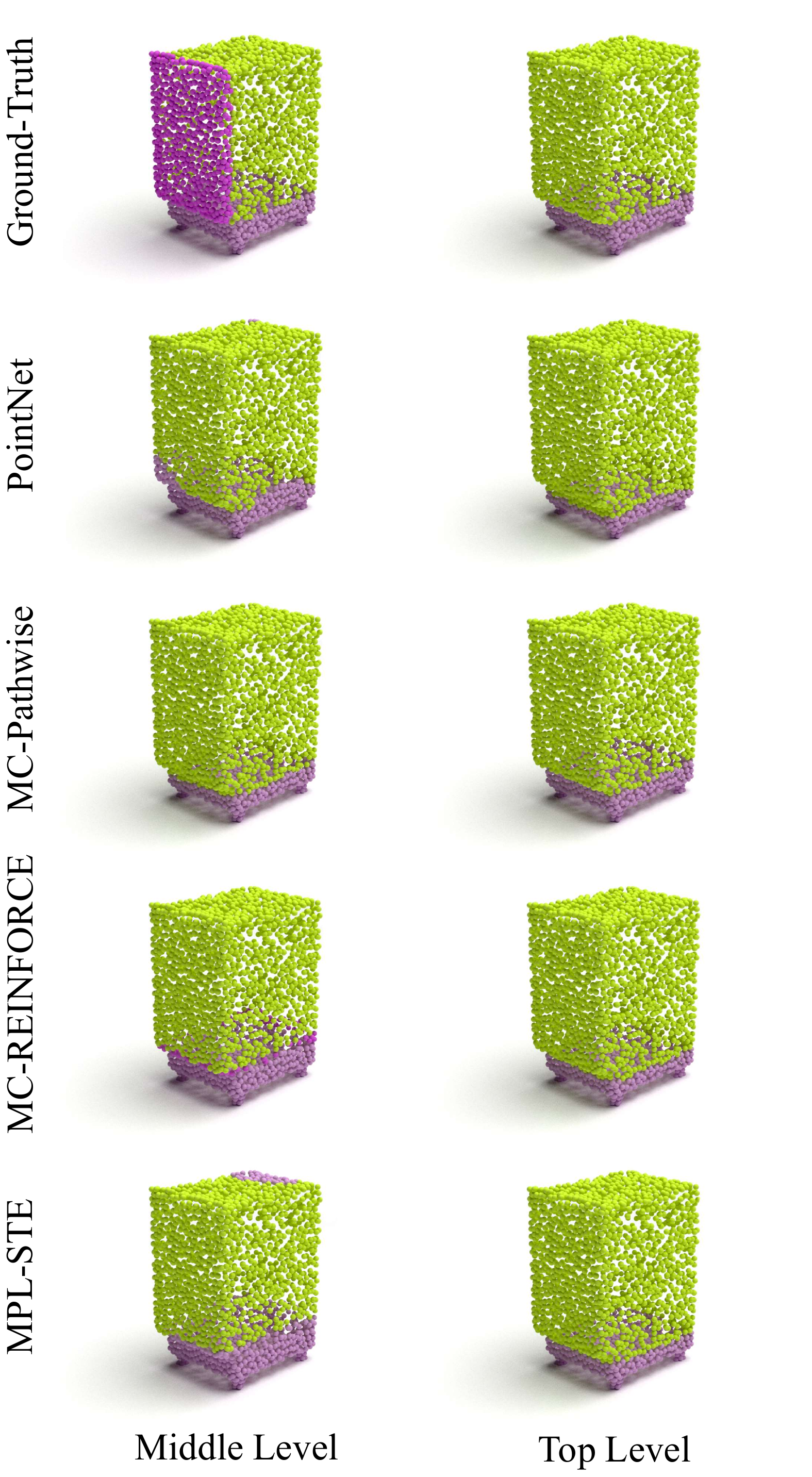}}
  \subfigure[Display]{\includegraphics[width=0.33\textwidth]{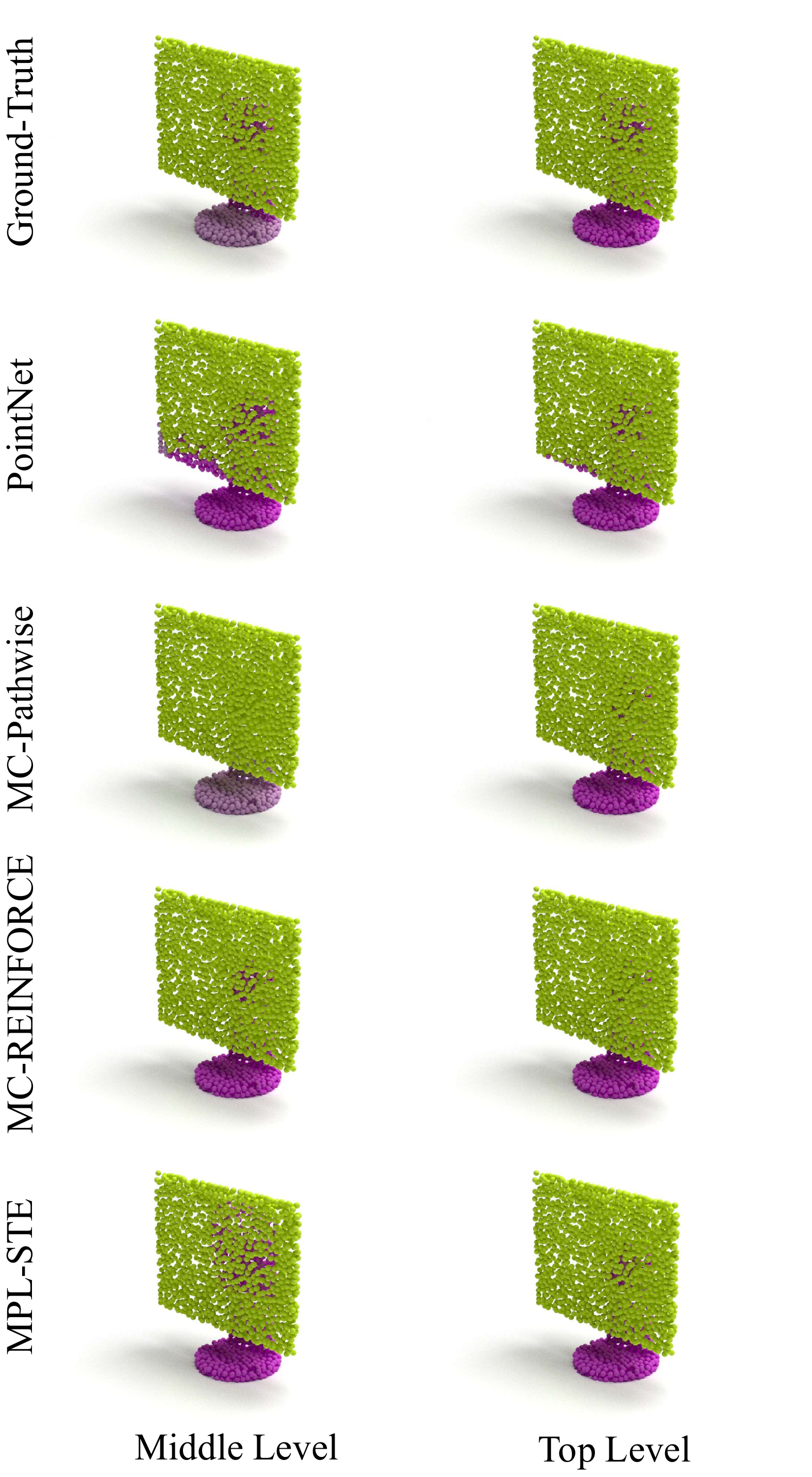}}
  \subfigure[Knife]{\includegraphics[width=0.33\textwidth]{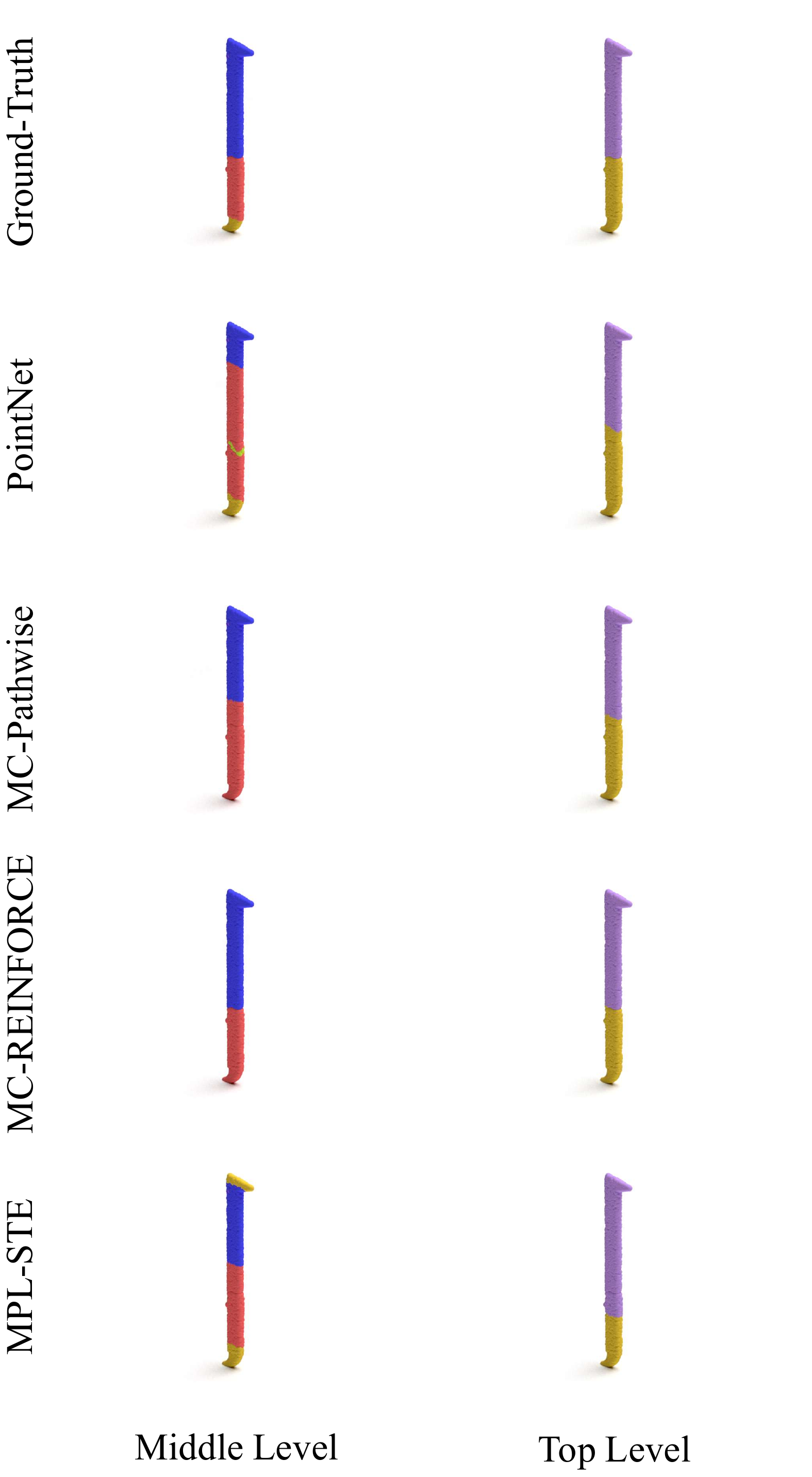}\label{subfig:knife}}
\end{figure*}
\begin{figure*}[htbp]
  \centering
  \subfigure[Storage]{\includegraphics[width=0.33\textwidth]{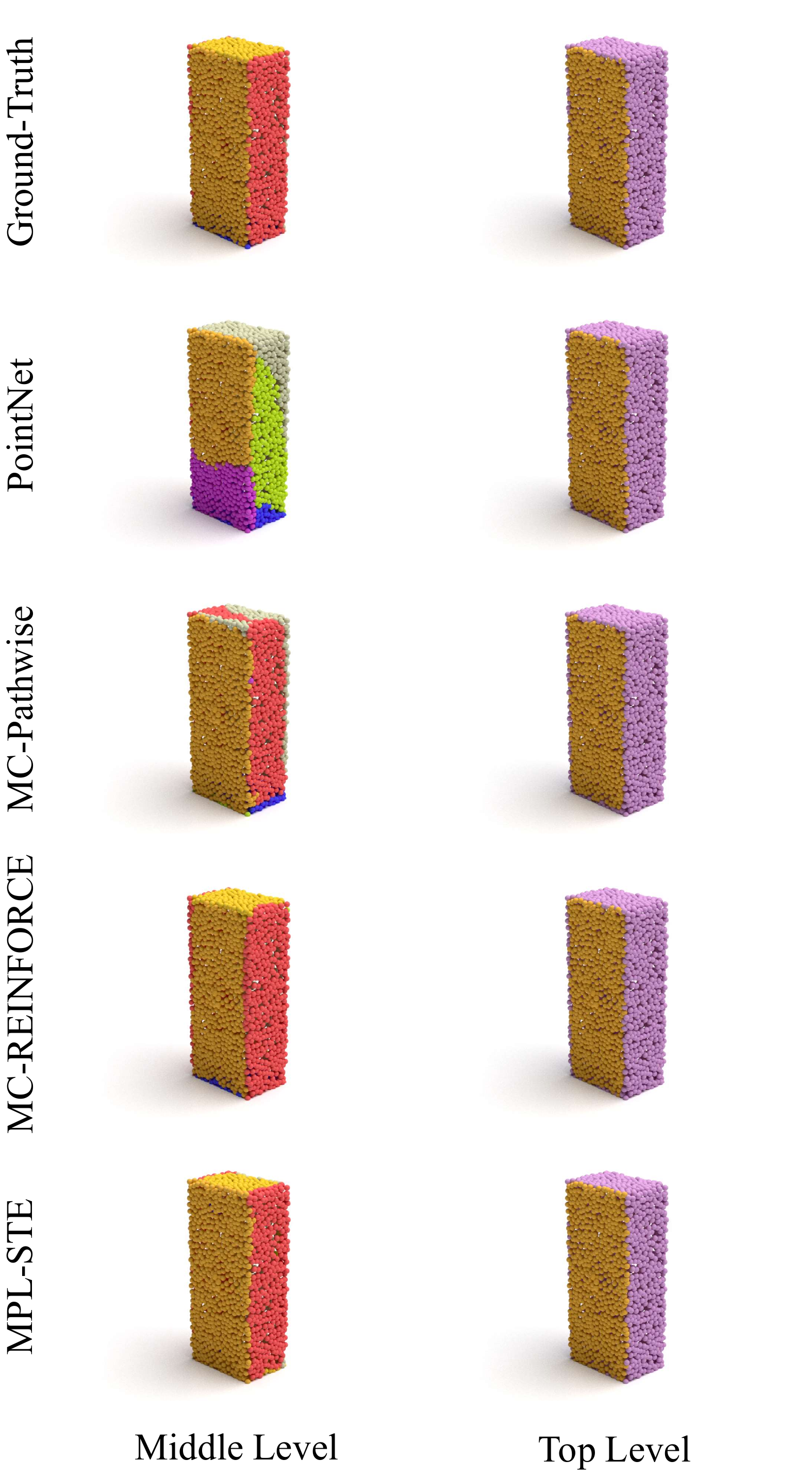}}
  \subfigure[Table]{\includegraphics[width=0.33\textwidth]{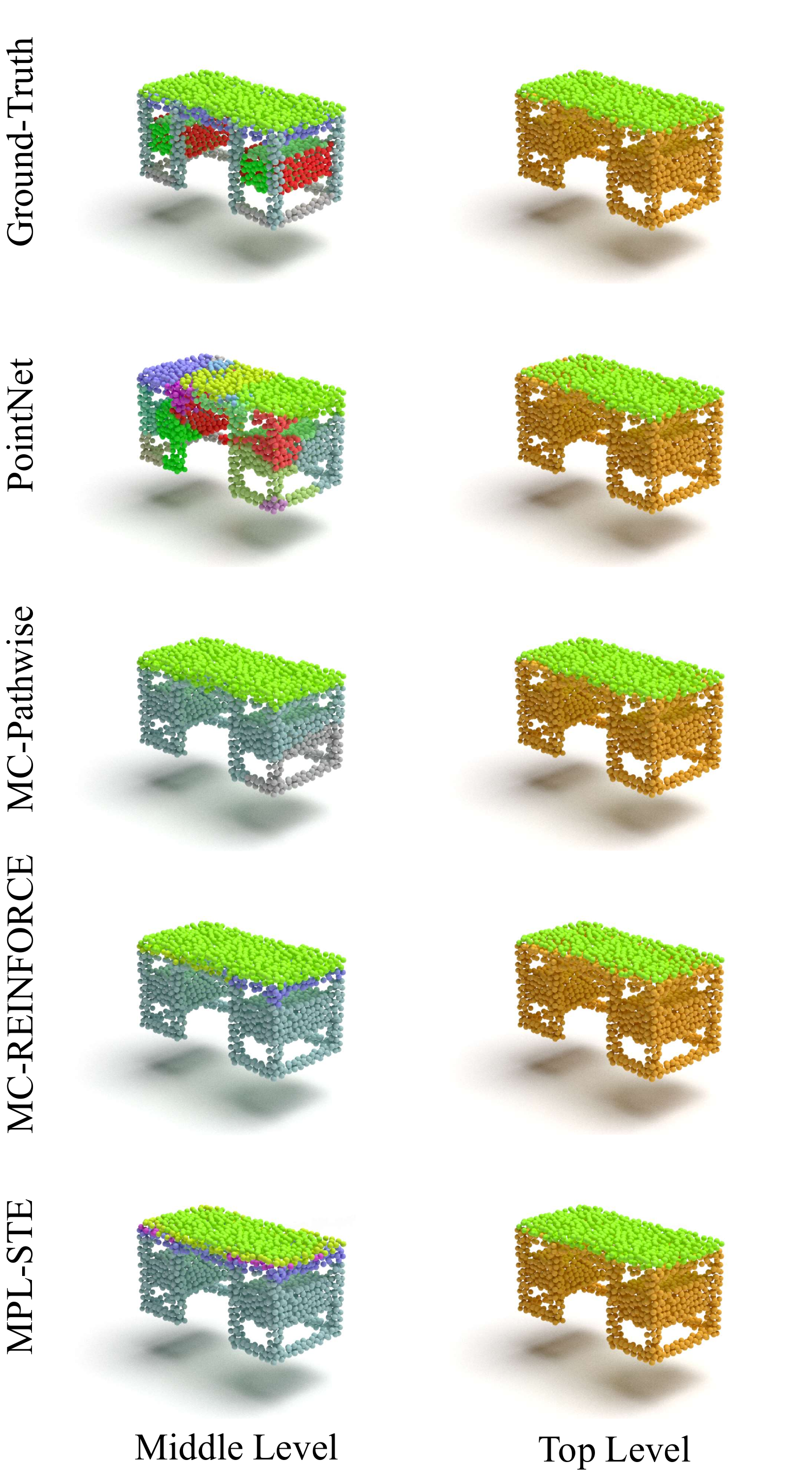}}
  \subfigure[Table]{\includegraphics[width=0.33\textwidth]{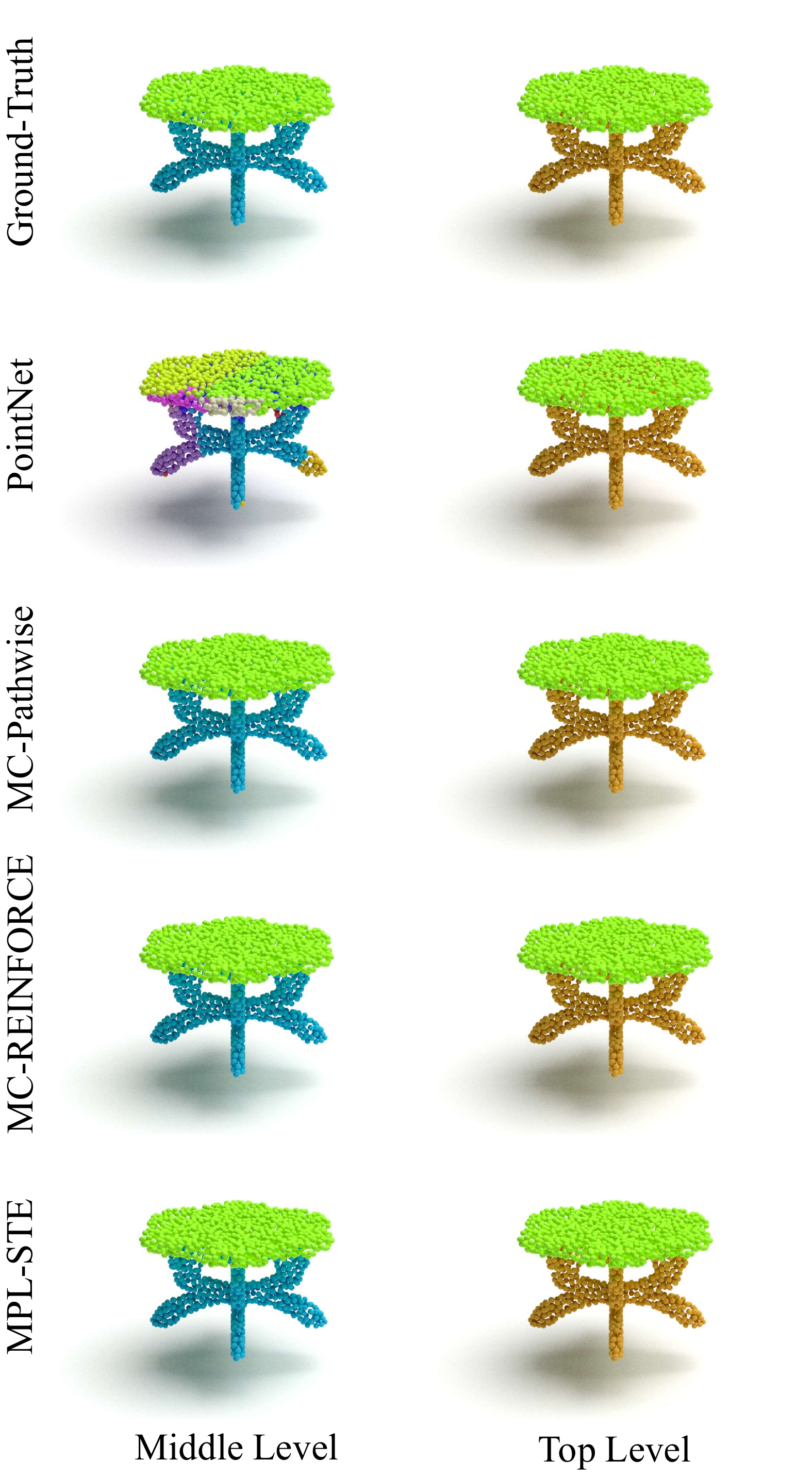}}
  \caption{Visual comparison of the point cloud segmentation.}
  \label{fig:vis_results}
\end{figure*}

\subsubsection{Benchmark Methods}
\label{subsubsec:benchmark}

Since few existing methods currently have the same capability and the same experimental setting as ours, we modify the existing competing methods to enable a fair comparison.
We choose state-of-the-art deep learning models for 3D point clouds as benchmark methods, including PointNet \cite{qi2017pointnet}, PointNet++ \cite{qi2017pointnet++}, PointCNN \cite{li2018pointcnn}, and PointConv \cite{wu2019pointconv}.
We carefully adjusted the size of individual models so that the number of parameters of each model is roughly the same or on the same magnitude as ours.
Among them, we use the labels at the top level as the supervised signal for training; for the unsupervised middle level, we select the intermediate output from a middle layer, \eg, the output before the feature transformation layer \cite{qi2017pointnet} of PointNet, 
and then perform the mini-batch K-Means algorithm \cite{sculley2010web} to cluster these features.
Finally, we perform the same matching process as before to ground the clustering labels to the latent part labels.
We evaluate their performances using the aforementioned metrics in Section~\ref{subsubsec:metrics}.

\subsubsection{Results}

Table \ref{tab:seg_results} and Table \ref{tab:mean_seg_acc} shows the results for the weakly-supervised segmentation task under the two approximated inference strategies, \ie, most-probable-latent inference with STE estimator (MPL-STE), Monte Carlo inference with REINFORCE estimator (MC-REINFORCE), and Monte Carlo inference with Pathwise estimator (MC-Pathwise).
We select the best model under different categories based on the highest average accuracy of the middle and top level on the testing set.
We report the mean accuracy across categories at middle and top levels in Table ~\ref{tab:mean_seg_acc}. We can see that at the unsupervised middle level, our MC-Pathwise, MPL-STE, and MC-REINFORCE bring 5.40\%, 6.88\%, and 7.16\% gain compared to the best performing baseline method PointNet, respectively. 
Similarly, our methods also bring an average of 2.0\% improvement at the supervised top level.

As we can see from accuracies of individual categories in Table \ref{tab:seg_results}, our proposed models achieve the best performance on most categories compared to other baseline methods.
Our model brings significant improvements over baseline methods at the unsupervised middle level in categories where the number of parts is large, \eg, \texttt{Chair} and \texttt{Table}.
This shows that the proposed model learns better part-whole hierarchies compared to the competing methods. 
In particular, MC-REINFORCE works the best on 7/16 and 6/16 categories at the middle and top levels respectively. 
MPL-STE works the best on 8/16 and 4/16 categories at the middle and top levels respectively, whereas MC-Pathwise works the best for 1/16 categories at both middle and top levels.
Overall, based on the average accuracy, MC-REINFORCE performs the best among all these variants.

\begin{figure*}[t]
  \centering
  \subfigure[Clock]{\includegraphics[width=0.33\textwidth]{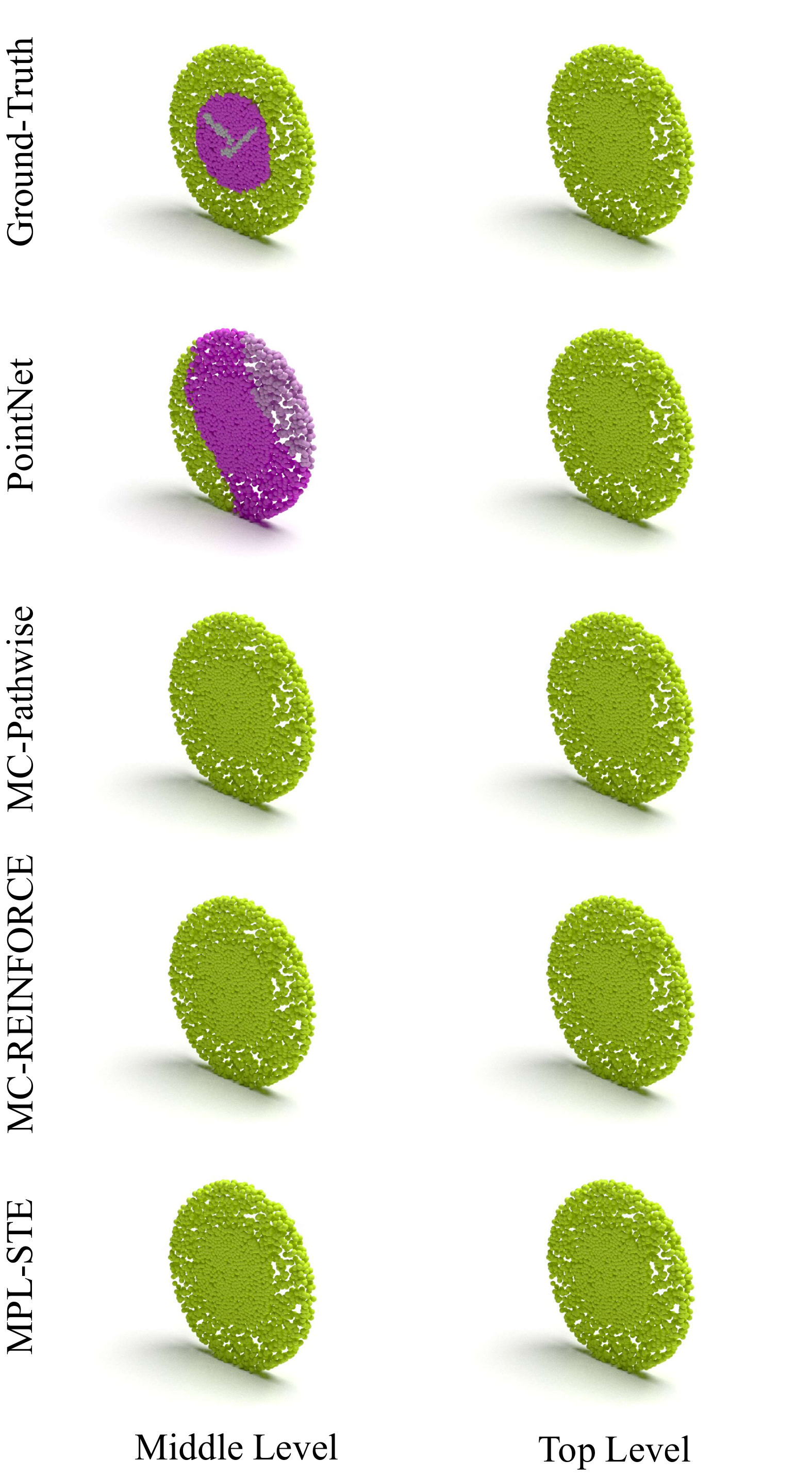}\label{subfig:fail_clock}}
  \subfigure[Storage]{\includegraphics[width=0.33\textwidth]{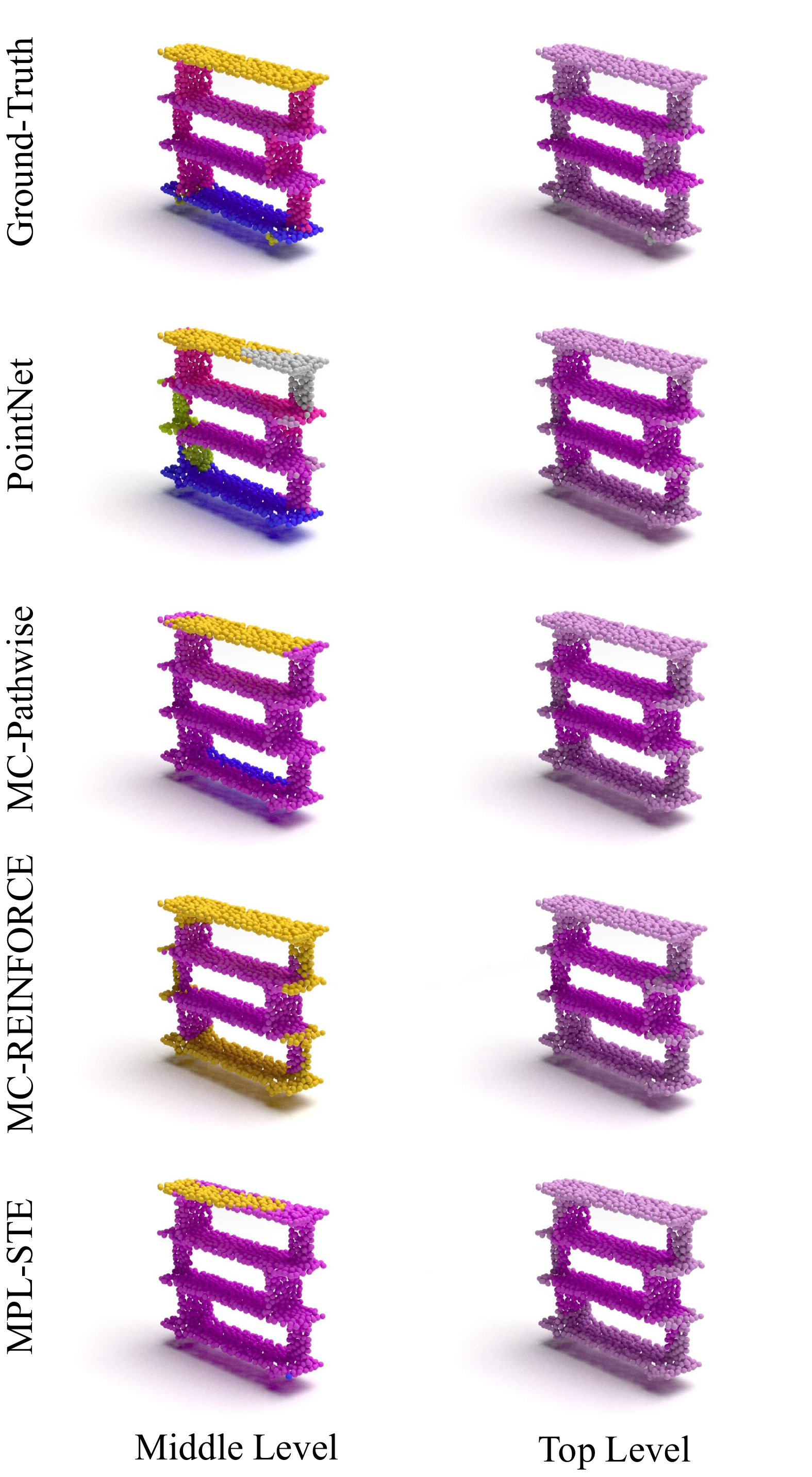}\label{subfig:fail_storage}}
  \subfigure[Table]{\includegraphics[width=0.33\textwidth]{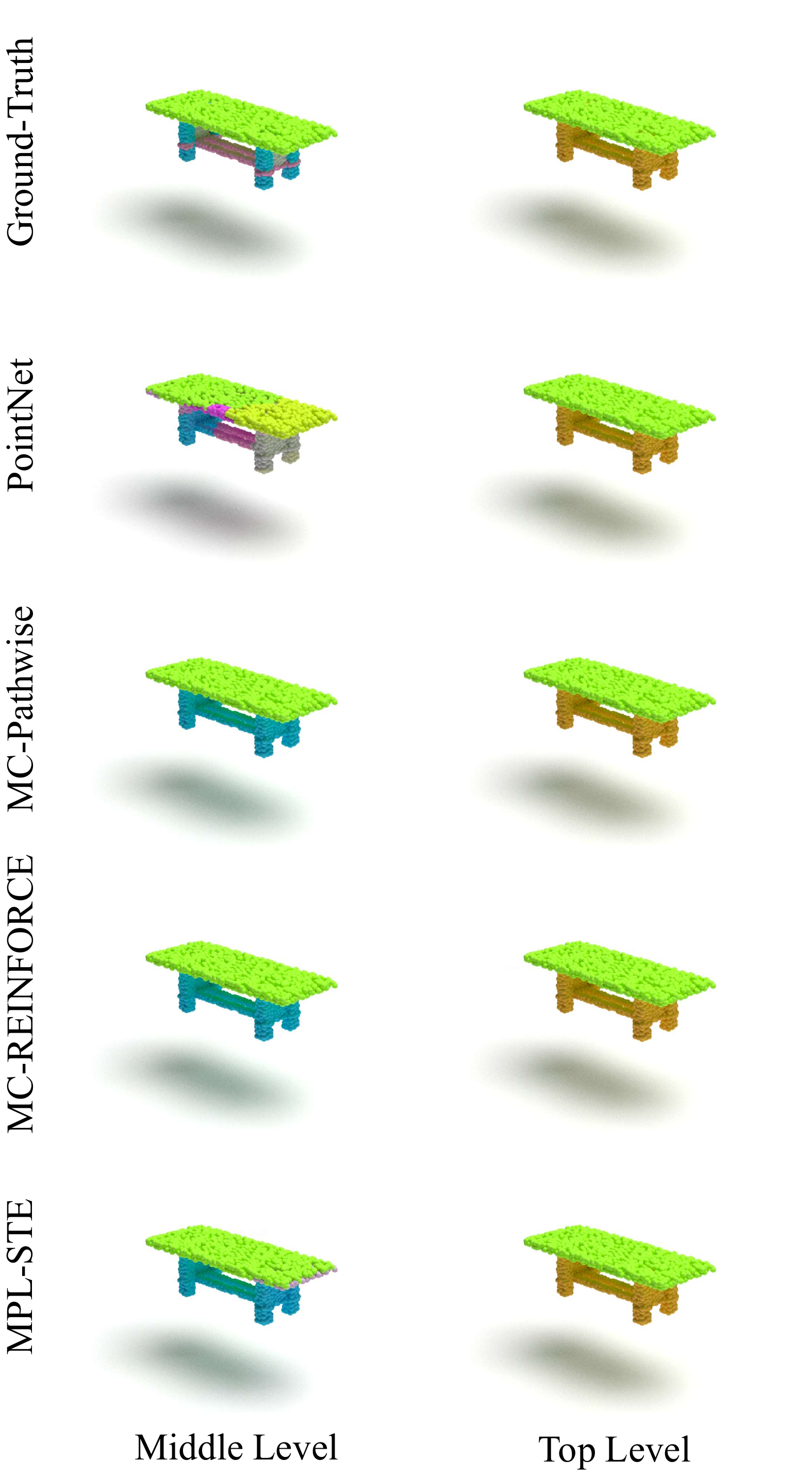}\label{subfig:fail_table}}
  \caption{Failure cases of the segmentation results.}
  \label{fig:failure_case}
\end{figure*}

\subsubsection{Visualization}

Fig.~\ref{fig:vis_results} visualizes the multi-level segmentation results of the proposed latent variable model with three gradient estimation methods compared to the strongest competitor PointNet \cite{qi2017pointnet}.
At the supervised top level, we can see that our proposed models is slightly better or comparable to PointNet on most categories, especially on the \texttt{Table} class.
In particular, our proposed MC-REINFORCE model has more accurate segmentation boundaries than PointNet on some categories, especially the cutting edge of the \texttt{Knife} in Fig.~\ref{subfig:knife}.
At the unsupervised middle level, our proposed models achieve better segmentation results than PointNet on most categories, especially on the challenging categories like \texttt{Chair} and \texttt{Table}.
This illustrates the effectiveness of our model on learning the latent parts within the part-whole hierarchy on 3D point clouds.

\subsubsection{Failure Case Analysis}

We show several failure cases of our models in Fig.~\ref{fig:failure_case}.
\begin{enumerate}
  \item \textbf{Over-smoothed bottom-up representations.} We notice that in Fig.~\ref{subfig:fail_clock}, the hands and dial of the clock at the middle level are not properly segmented. This is partly due to the fact that normal vectors of all points are very similar. Our models may learn over-smoothed representations at the middle level even though the different parts are distinguishable. 
  \item \textbf{Lack of top-down guidance.} As illustrated in Fig.~\ref{subfig:fail_storage}, we see that at the middle level, our model sometimes misclassifies the boards at the top and bottom of the shelf, though they exhibit similar structures while belonging to different categories. Similarly, in Fig.~\ref{subfig:fail_table}, we see that our models only segment the table into two or three sub-parts despite that there are are way more sub-parts annotated in the ground truth. We argue that such cases are too difficult to distinguish solely based on bottom-up information as these sub-parts are structurally very similar. More top-down guidance or prior information is required to inform the model during learning.
\end{enumerate}

\subsection{Ablation Studies}

In this section, we conduct ablation studies on the important hyperparameters of our models.

\subsubsection{Sampling on Inference}

\begin{figure*}[t]
  \centering
  \subfigure[Chair (MC-REINFORCE)]{\includegraphics[width=0.33\textwidth]{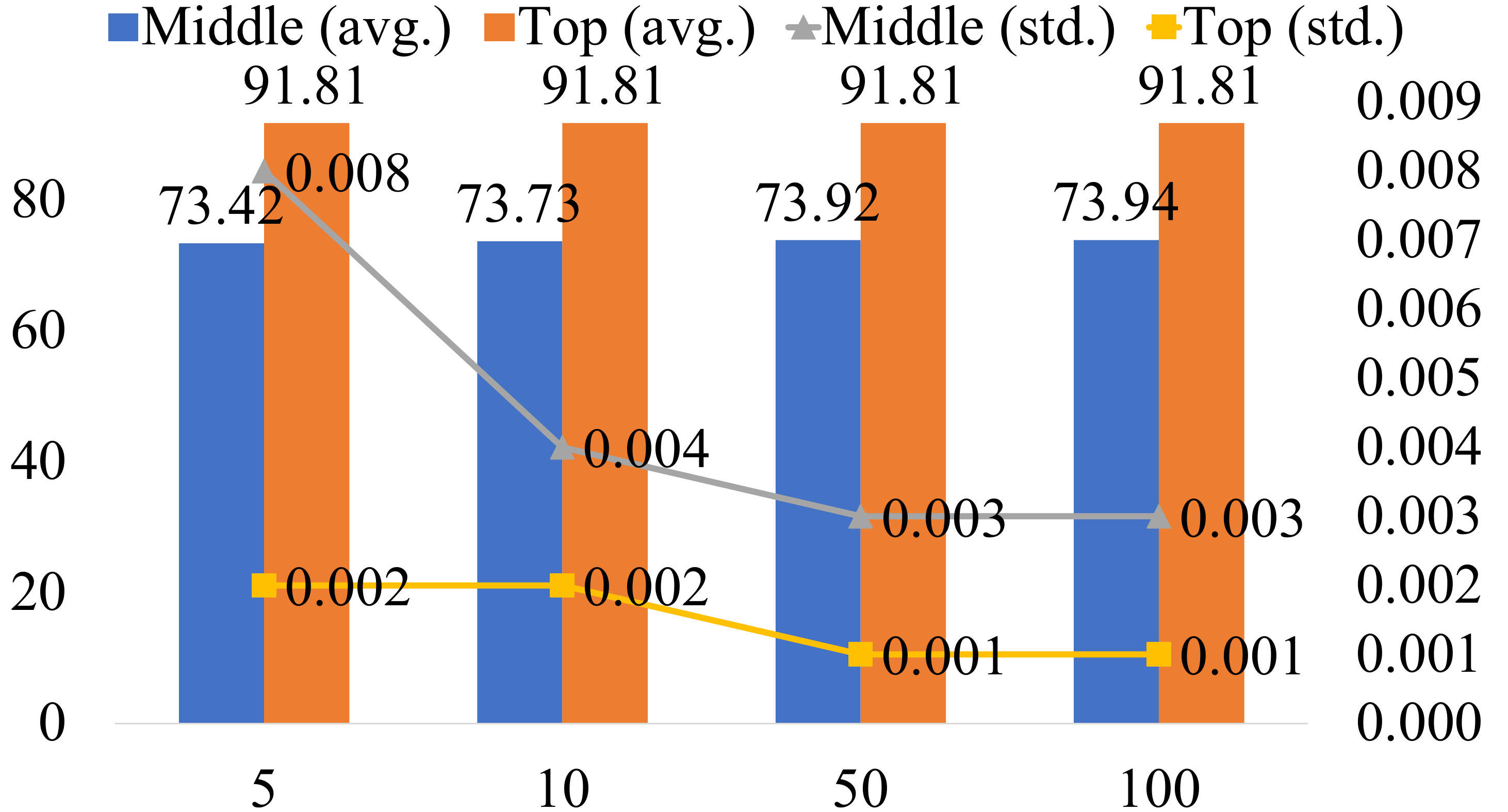}}
  \subfigure[Lamp (MC-REINFORCE)]{\includegraphics[width=0.33\textwidth]{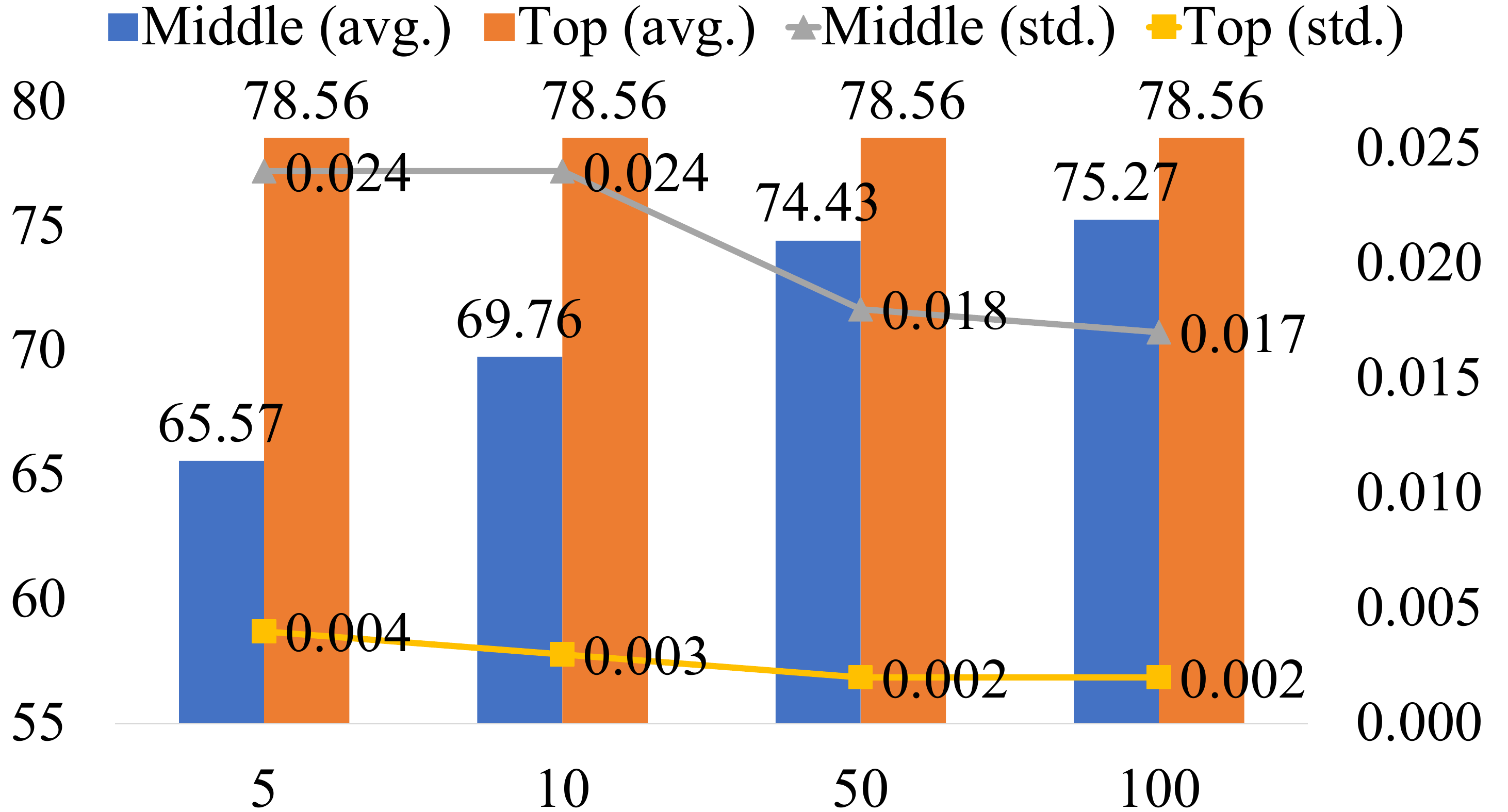}}
  \subfigure[Tabel (MC-REINFORCE)]{\includegraphics[width=0.33\textwidth]{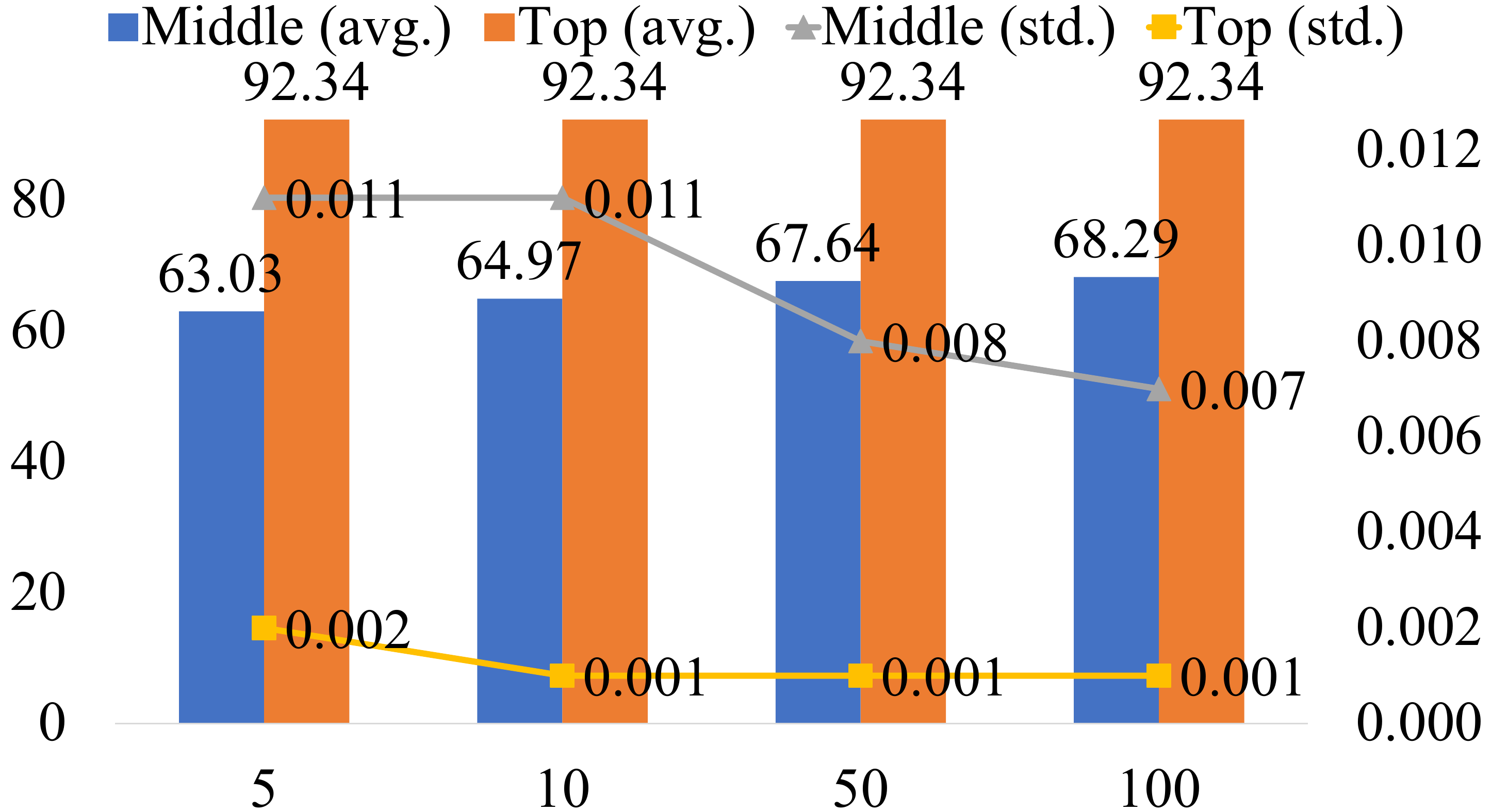}}
  \subfigure[Chair (MC-Pathwise)]{\includegraphics[width=0.33\textwidth]{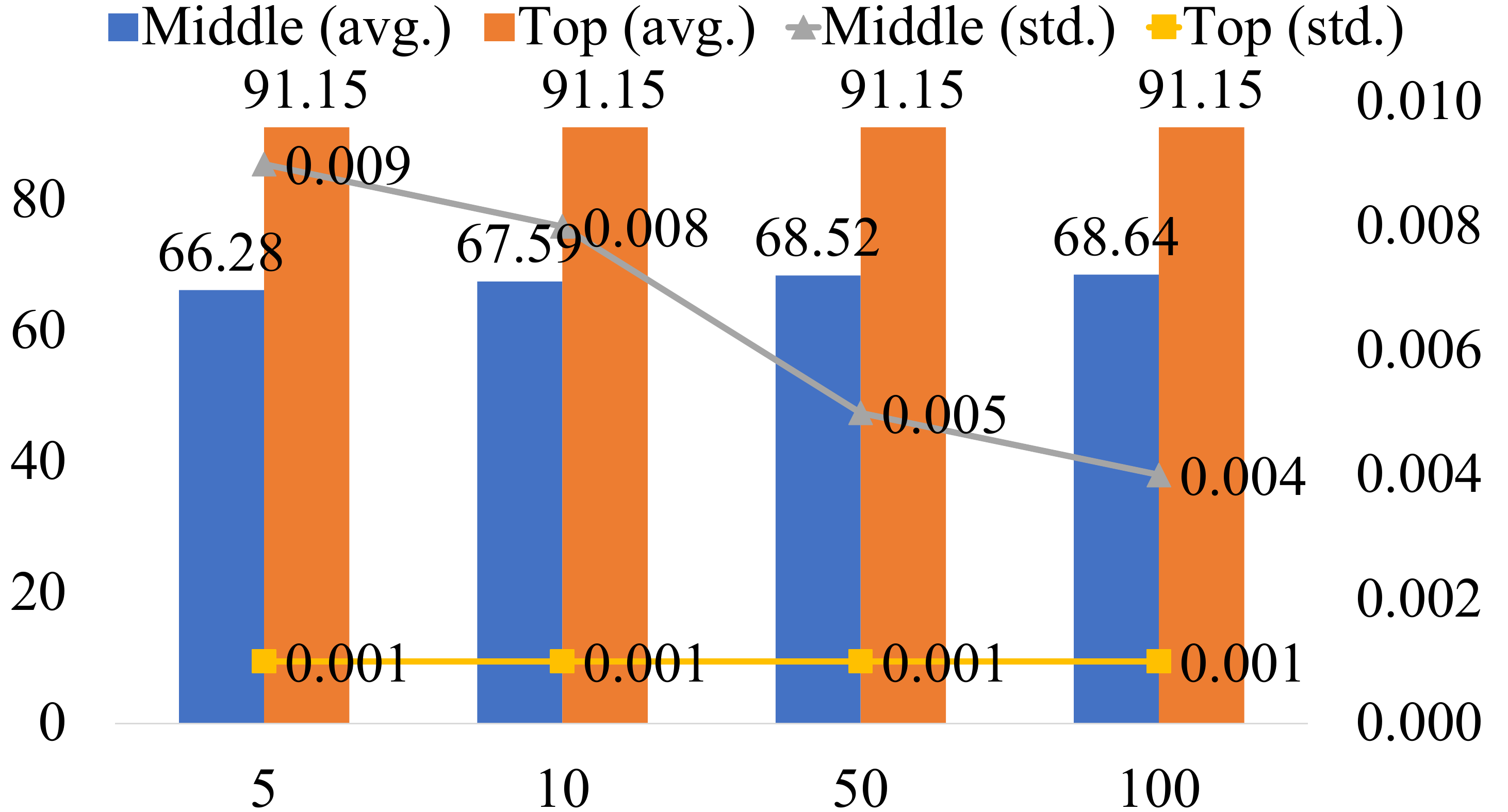}}
  \subfigure[Lamp (MC-Pathwise)]{\includegraphics[width=0.33\textwidth]{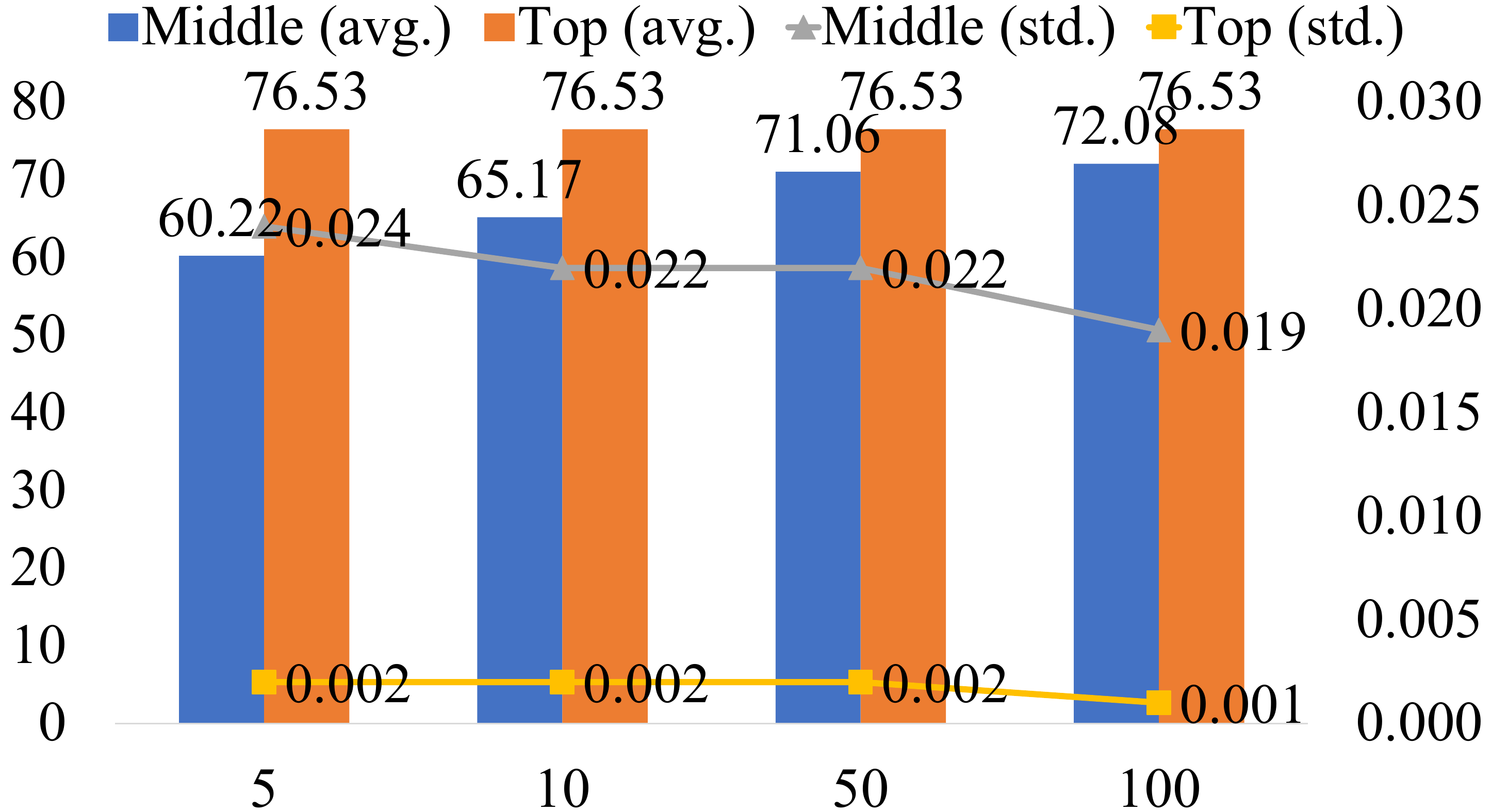}}
  \subfigure[Tabel (MC-Pathwise)]{\includegraphics[width=0.33\textwidth]{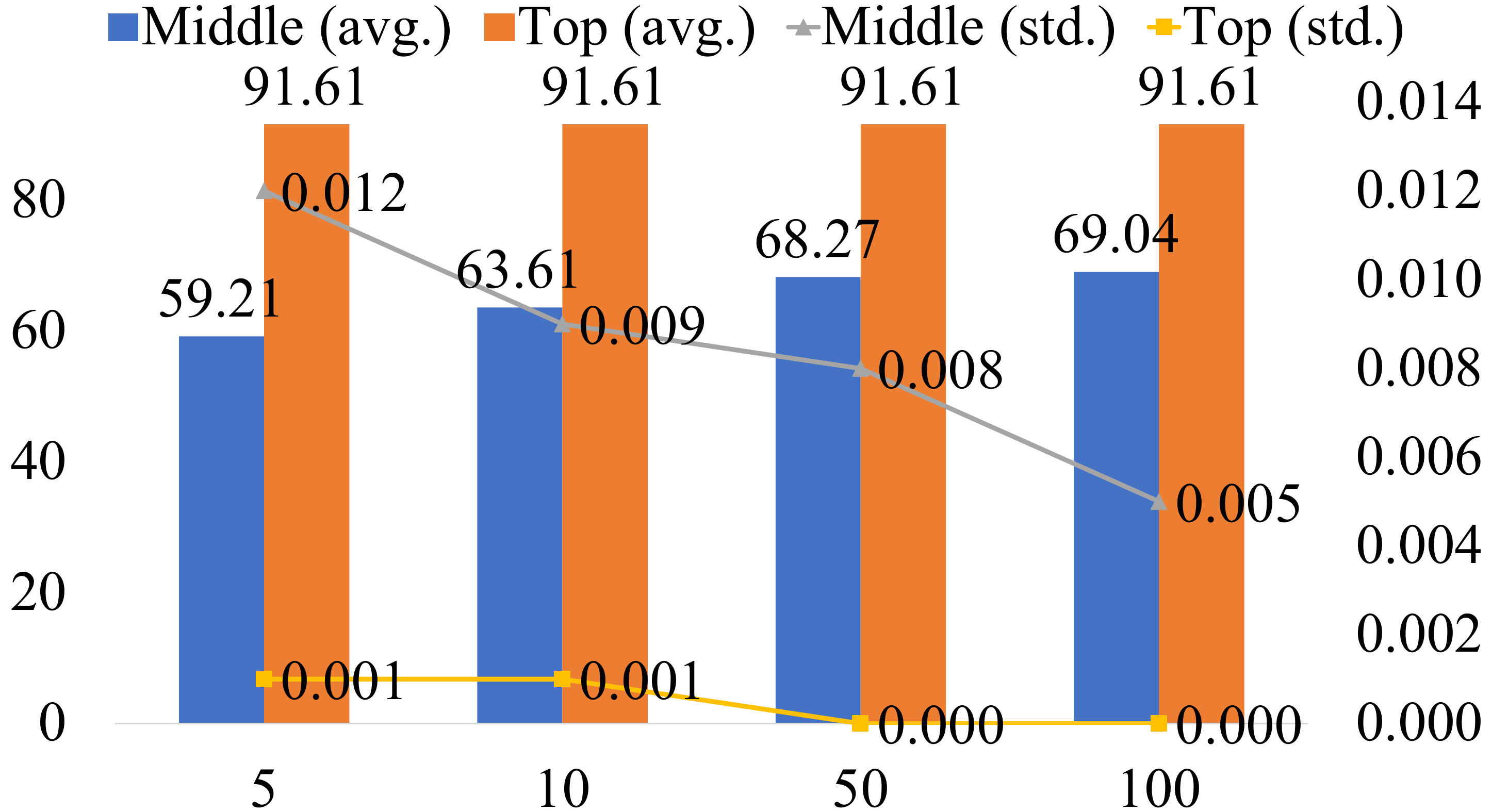}}
  \caption{\textbf{Segmentation results on the Monte Carlo inference.} The horizontal axis is the number of samples, and the left vertical axis and the right vertical axis represent the average segmentation accuracy (avg., \%) and the standard deviation (std., \%), respectively.}
  \label{fig:sampling_on_inference}
\end{figure*}

We first evaluate the Monte Carlo inference methods with REINFORCE and Pathwise gradient estimators.
As shown in Fig.~\ref{fig:sampling_on_inference}, we draw $\{5,10,50,100\}$ samples from the learned distribution $p_{\phi}(\z|\x)$ of the latent variable, and then average these samples to obtain the middle-level label $\z$; we then feed these samples to the decoder, and further estimate the label $\y$ of the top level by averaging the outputs of the decoder.
We repeat the above experiments $10$ times and report the average segmentation accuracy and standard deviation.
As we can see, the segmentation accuracies of the middle level are usually lower when the number of samples is relatively small (\eg, 5) since a small number of samples is insufficient to accurately characterize the probability distribution of the latent variable; while accuracies at the top level are higher since the learning benefits from the supervised objective.
Further, it is unsurprising that the variance of the performances at both the middle and top levels is generally large when we use a small number of samples.
Our models will obtain more stable results with more samples.
  
\subsubsection{Temperature Parameter \texorpdfstring{$\tau$}{tau}}
\label{subsubsec:tau_ablation}

\begin{figure}[htbp]
  \centering
  \subfigure[Middle Level]{\includegraphics[width=0.9\columnwidth]{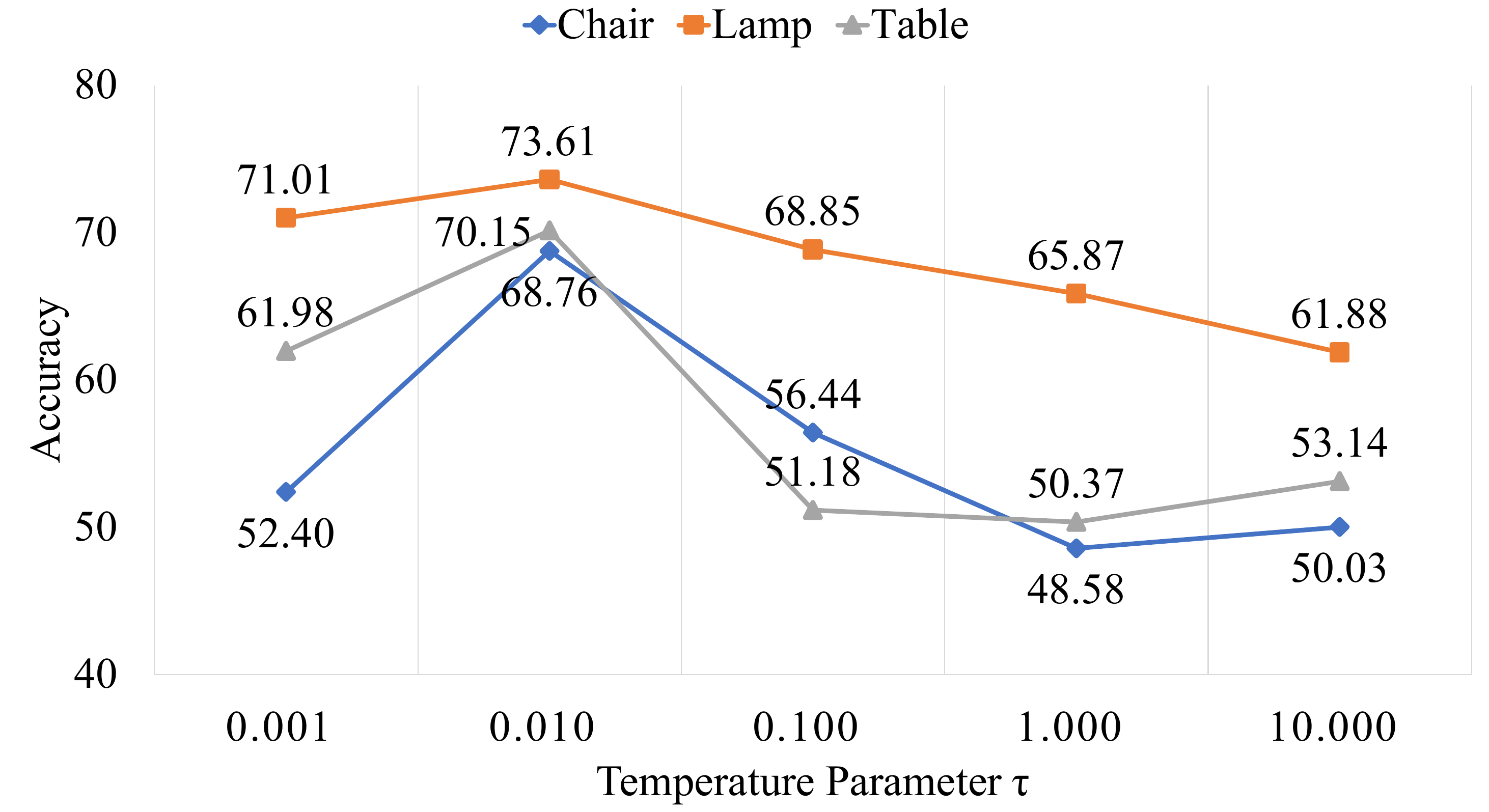}}
  \subfigure[Top Level]{\includegraphics[width=0.9\columnwidth]{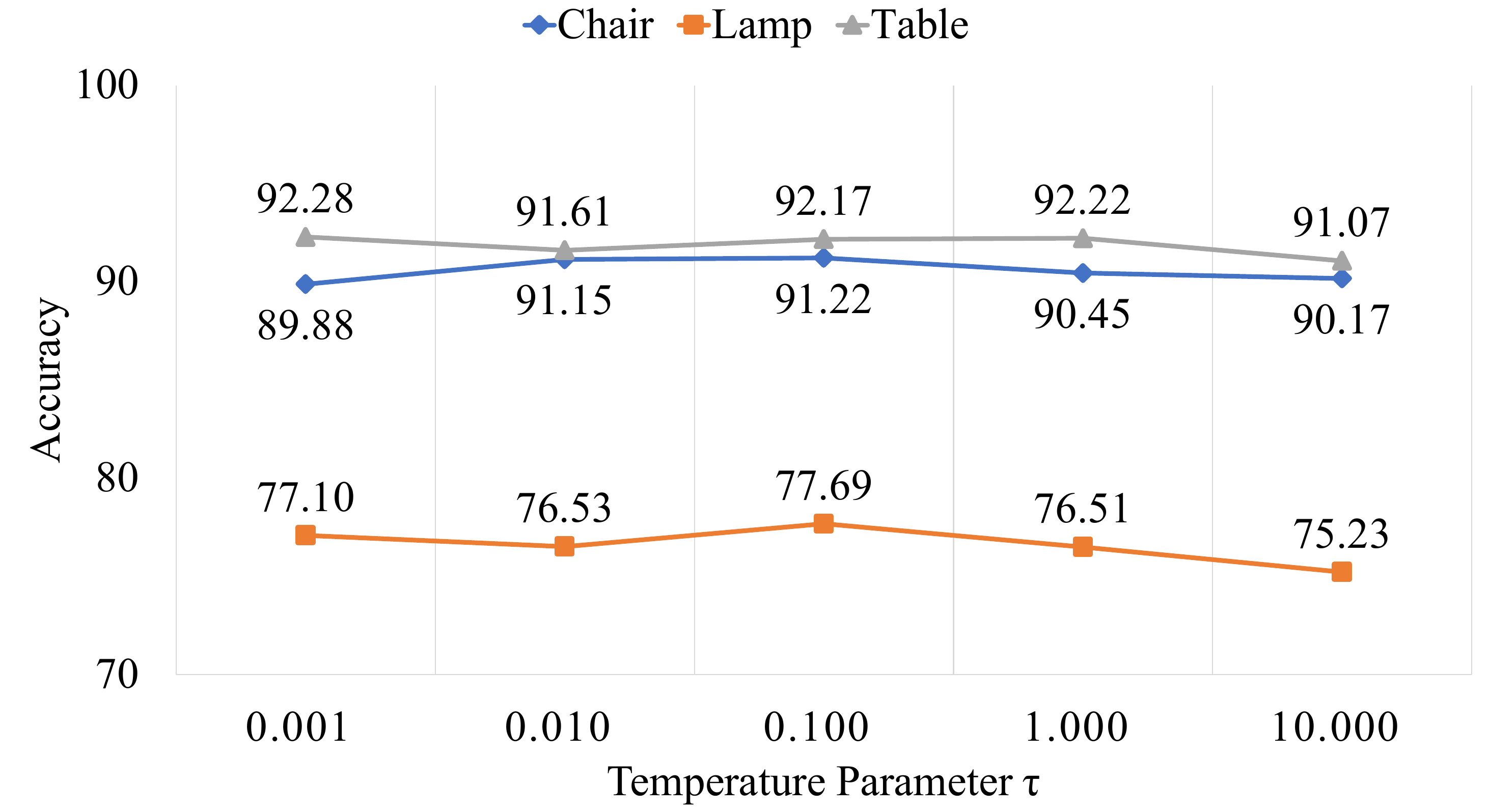}}
  \caption{Segmentation accuracy of our model with different temperature parameter $\tau$.}
  \label{fig:different_tau}
\end{figure}

We next evaluate the effect of the temperature $\tau$ of the pathwise gradient estimator on the segmentation results.
We adopt five different values of $\tau$ in $\{0.001,0.01,0.1,1.0,10.0\}$ to train our model.
Fig.~\ref{fig:different_tau} presents the segmentation accuracies of five different temperature values on three categories \texttt{Chair}, \texttt{Lamp} and \texttt{Table}.
At the middle level, we see that when $\tau$ is set to a large value, our model is able to produce reasonable results.
The performance keeps improving with $\tau$ decreasing in general and achieves the state-of-the-art, and finally drops as $\tau$ decreases further.
At the top level, we see that the performance does not change much due to the supervised training.

\begin{table}[htbp]
  \caption{Segmentation accuracy (\%) comparison with or without control variate on the REINFORCE estimator.}
  \label{tab:ablation_cv}
  \begin{tabularx}{\columnwidth}{c|YY|YY}
  \hline
  \multirow{2}{*}{} & \multicolumn{2}{c|}{w/ Control Variate} & \multicolumn{2}{c}{w/o Control Variate} \\ \cline{2-5} 
   & Mid Level & Top Level & Mid Level & Top Level \\ \hline
  Chair & $73.75 \pm 0.19$ & $91.59 \pm 0.12$ & $72.39 \pm 0.95$ & $91.50 \pm 0.12$ \\
  Lamp & $76.23 \pm 0.52$ & $77.49 \pm 0.59$ & $75.65 \pm 1.86$ & $76.78 \pm 0.82$ \\
  Table & $71.64 \pm 0.27$ & $92.33 \pm 0.07$ & $70.40 \pm 0.58$ & $92.00 \pm 0.21$ \\ \hline
  \end{tabularx}
\end{table}

\subsubsection{Control Variate}

We further evaluate the effect of the control variate technique in the REINFORCE estimator.
We train five models separately using the REINFORCE estimator with and without control variate on three categories \texttt{Chair}, \texttt{Lamp}, and \texttt{Table}, and report the average accuracy and the standard deviation.
As shown in Table~\ref{tab:ablation_cv}, the segmentation results with control variate are better than those without control variate on all three categories, and the variance is significantly lower with the control variate.
The better average accuracies are likely contributed by the low-variance gradient estimation in learning.

\section{Discussion and Conclusion}
\label{sec:conclusion}
\subsection{Discussion}
\label{subsec:discuss}

\textbf{Generalization to more levels.}
Our proposed latent variable model can be generalized to multiple levels in a straightforward manner.
From Eq.~(\ref{eq:latent_model}), we can further decompose the conditional distribution of the latent variable as,
\begin{align}\label{eq:level_generalization}
  p(\y\vert\x) &= \sum_{\z_T}p_{\theta_T}(\y\vert\z_T,\x) p_{\phi_T}(\z_T\vert\x), \nonumber \\
  \text{with} \quad p_{\phi_T}(\z_T \vert\x) &= \sum_{\z_{T-1}}p_{\theta_{T-1}}(\z_T \vert\z_{T-1},\x) p_{\phi_{T-1}}(\z_{T-1}\vert\x), \nonumber \\
  &\vdots \nonumber \\
  p_{\phi_{2}}(\z_{2}\vert\x) &= \sum_{\z_{1}}p_{\theta_{1}}(\z_{2}\vert\z_{1},\x) p_{\phi_{1}}(\z_{1}\vert\x),
\end{align}
where the conditional distribution of the classifier $p(\y\vert\x)$ is decomposed into $T+1$ levels: the latent class label at each level is encoded by the latent variable $\z_1,...,\z_T$.
Eq.~(\ref{eq:level_generalization}) assumes the interaction between adjacent two levels.
More complicated interaction among different levels could be explored in the future.

\noindent \textbf{Limitations.}
Since the proposed model is weakly supervised, the representation learning of the unsupervised middle level may be strongly influenced by the supervised top level in the absence of prior information.
Hence, the segmentation results of the middle level may tend to be similar to ones at the top level.
Also, our proposed model requires Monte Carlo sampling during training, which can be inefficient when the number of levels is large.

\subsection{Conclusion}
\label{subsec:conclude}
In this paper, we propose a novel latent variable model for learning part-whole hierarchies on 3D point clouds in a weakly-supervised way.
We design an encoder-decoder framework that explicitly captures part-whole hierarchies of point clouds.
We explore two simple yet effective approximated inference algorithms and three gradient estimation methods for learning discrete latent variables.
Experimental results show that our models achieve the state-of-the-art performance on weakly-supervised multi-level point cloud segmentation.
In the future, we plan to generalize our model to more levels, and apply to more applications such as weakly supervised point cloud classification.

\bibliographystyle{IEEEtran}

\end{document}